\documentclass{article} 
\usepackage{iclr2026_conference,times}

\usepackage{amsmath,amsfonts,bm}

\def\eqref#1{equation~\ref{#1}}

\def\1{\bm{1}}

\DeclareMathAlphabet{\mathsfit}{\encodingdefault}{\sfdefault}{m}{sl}
\SetMathAlphabet{\mathsfit}{bold}{\encodingdefault}{\sfdefault}{bx}{n}

\usepackage{hyperref}
\usepackage{url}

\usepackage{graphicx}
\usepackage{wrapfig}
\usepackage{pifont}
\usepackage{dsfont}
\usepackage{comment}
\usepackage{amsmath}    
\usepackage{mathrsfs}
\usepackage{booktabs}
\usepackage[table]{xcolor}
\usepackage{tabu}
\usepackage{multirow}
\usepackage{mathrsfs}
\usepackage{xspace}
\makeatletter
\DeclareRobustCommand\onedot{\futurelet\@let@token\@onedot}
\def\@onedot{\ifx\@let@token.\else.\null\fi\xspace}

\def\eg{\emph{e.g}\onedot} 
\def\ie{\emph{i.e}\onedot}

\makeatother

\title{FreeViS: Training-free Video Stylization with Inconsistent References}

\author{Jiacong Xu$^1$\quad
Yiqun Mei$^2$\quad
Ke Zhang$^1$\quad
Vishal M. Patel$^1$\\
$^1$ Johns Hopkins University\quad$^2$ Adobe Research\\
\texttt{\{jxu155, kzhang99, vpatel36\}@jhu.edu
ymei@adobe.com} \\
}

\iclrfinalcopy 
\begin{document}

\maketitle

\begin{figure}[ht]
    \vspace{-1ex}
    \includegraphics[width=\linewidth]{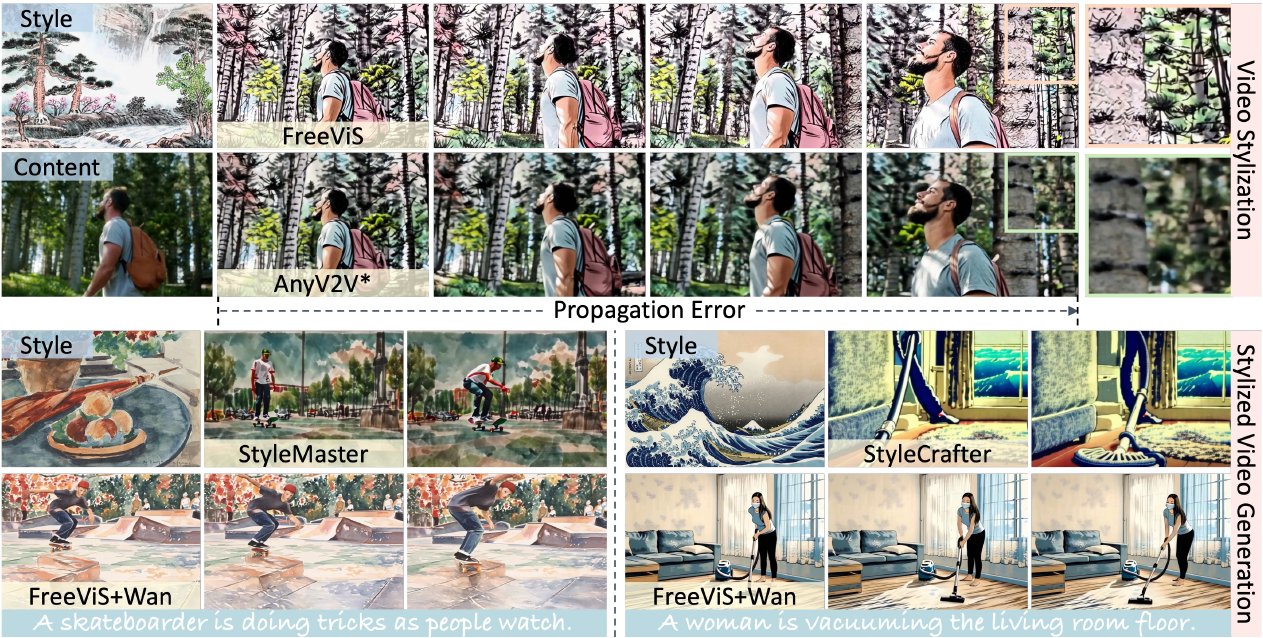}
    \vspace{-4ex}
    \caption{
    Previous works (e.g., AnyV2V \cite{ku2024anyv2v}) suffer from propagation errors inherent to their single-reference inputs. Combined with a text-to-video model \cite{wan2025wan}, FreeViS outperforms existing methods \cite{ye2025stylemaster,liu2024stylecrafter} on stylized video generation.
}
    \label{fig:teaser}
    \vspace{-1ex}
\end{figure}

\begin{abstract}
Video stylization plays a key role in content creation, but it remains a challenging problem. Naïvely applying image stylization frame-by-frame hurts temporal consistency and reduces style richness. Alternatively, training a dedicated video stylization model typically requires paired video data and is computationally expensive. In this paper, we propose FreeViS, a training-free video stylization framework that generates stylized videos with rich style details and strong temporal coherence. Our method integrates multiple stylized references to a pretrained image-to-video (I2V) model, effectively mitigating the propagation errors observed in prior works, without introducing flickers and stutters. In addition, it leverages high-frequency compensation to constrain the content layout and motion, together with flow-based motion cues to preserve style textures in low-saliency regions. Through extensive evaluations, FreeViS delivers higher stylization fidelity and superior temporal consistency, outperforming recent baselines and achieving strong human preference. Our training-free pipeline offers a practical and economic solution for high-quality, temporally coherent video stylization. The code and videos can be accessed via \url{https://xujiacong.github.io/FreeViS/}
\end{abstract}

\section{Introduction}
Style transfer has been a long-standing research topic in computer vision, with applications spanning art, education, advertising, and entertainment. In recent years, tremendous progress has been made in image style transfer \cite{gatys2016image, huang2017arbitrary, liu2021adaattn, chung2024style, xu2025stylessp} , driven by advances in deep learning architectures and generative modeling. While the focus has been on still images, videos have become a more popular medium for everyday communication in recent years. Yet video stylization remains significantly underexplored compared to its image-based counterpart. As content distribution changes across time, naïvely applying image stylization frame by frame leads to severe flickering artifacts, resulting in poor temporal consistency. Even though advanced temporal smoothing techniques \cite{duan2023fastblend} can mitigate flickering artifacts, they often do so at the expense of losing style richness, resulting in overly smoothed textures and reduced visual plausibility.

Previous video stylization works \cite{wang2020consistent, pande2023vist3d} have attempted to improve temporal consistency by modifying convolutional architectures. However, their quality remains inferior to that of recent diffusion-based image stylization approaches. More recently, several diffusion-based methods \cite{liu2024stylecrafter, ye2025stylemaster} have been proposed for stylized text-to-video (T2V) generation, but these approaches are not directly applicable to video-to-video (V2V) editing. In addition, our experiments also found that neither recent unified video editing frameworks \cite{jiang2025vace} nor text-driven editing methods \cite{geyer2024tokenflow} are able to perfectly transfer style across videos. Beyond these works, directly fine-tuning a video diffusion model (\eg, DiT-based architectures) is also impractical, as it demands large-scale original–stylized video pairs, which are difficult to obtain. Simply adopting the perceptual training strategy, as in image stylization \cite{liu2021adaattn}, cannot ensure temporal consistency across frames. Moreover, substantial computational resources are required for fully finetuning.

On the other hand, reference-based video editing methods, such as AnyV2V \cite{ku2024anyv2v}, can leverage image stylization models to stylize the first frame, and then propagate it to the rest of the frames using pretrained image-to-video (I2V) models. In this way, no training or only limited training \cite{ouyang2024i2vedit} is required. However, due to the absence of stylized videos in the training phase of the base I2V model, it fails to properly handle the reference frame as it is out of the training distribution (see Appendix \ref{app:noise}), thus incapable of parsing and propagating style patterns from the first edited frame. As Figure \ref{fig:teaser} shows, these reference-based methods struggle to transfer style patterns to novel content in subsequent frames, resulting in a pronounced \textit{propagation error}. 

In this paper, we aim to tackle video stylization in a training-free manner, \ie by inverting a diffusion model, to achieve comprehensive style transfer while preserving temporal coherence. An intuitive solution to address \textit{propagation error} is to provide the base model with multiple reference frames across the entire video. However,
naïvely concatenating additional reference frames with noise latent leads to severe flickering and stuttering artifacts in the results. To address this, we introduce isolated attention with a carefully designed masking strategy to mitigate stylization inconsistencies, together with a novel approach to inject temporal dynamics into the reference latents. We also found that the inverted noises alone are insufficient to accurately reconstruct original video content, often missing structures in stylized results. To this end, we propose a strategy to extract high-frequency components from the reconstruction compensation of PnP inversion \cite{ju2024pnp} to compensate and constrain spatial layouts and motion trajectories.
Lastly, in plain regions with few salient features, style textures may vanish under large camera motion. To overcome this, we propose a method that leverages optical flow to constrain diffused attention areas in associated frames.

The contributions of this paper are summarized as follows: (1) We propose a novel reference-based and training-free framework: FreeViS, for video stylization, which effectively addresses the propagation error inherent in existing methods; (2) We conduct extensive experiments demonstrating the effectiveness of our design and showing that FreeViS achieves higher stylization quality compared to prior approaches; (3) We show that, when combined with existing T2V models, FreeViS also achieves stylized T2V generation with performance comparable to recent training-based methods.
\vspace{-2ex}

\section{Observations}

\subsection{Frequency Analysis of Intermediate Latents}
\label{sec:frequency}
Since the inverted noises alone cannot accurately reconstruct the videos (Appendix \ref{app:noise}),
we adopt PnP Inversion \cite{ju2024pnp} to progressively align the denoising trajectory with the inversion path. Building on this framework, we investigate the influence of low-frequency (LF) and high-frequency (HF) components of intermediate latents along the denoising process (Appendix \ref{app:frequency}). Recent image editing studies \cite{yang2023diffusion, xu2025stylessp} have shown that the HF components of the initial noise primarily determine the spatial layout of the generated images.
Here, we observe a similar phenomenon in I2V models: \textit{LF latents primarily govern the appearance and color distribution of the generated video, whereas HF components encode layout and motion cues.}

\vspace{-1ex}
\subsection{Cross-frame Attention}
\label{sec:cross_frame}
\begin{wrapfigure}{r}{0.6\textwidth}  
\vspace{-2ex}  
\includegraphics[width=\linewidth]{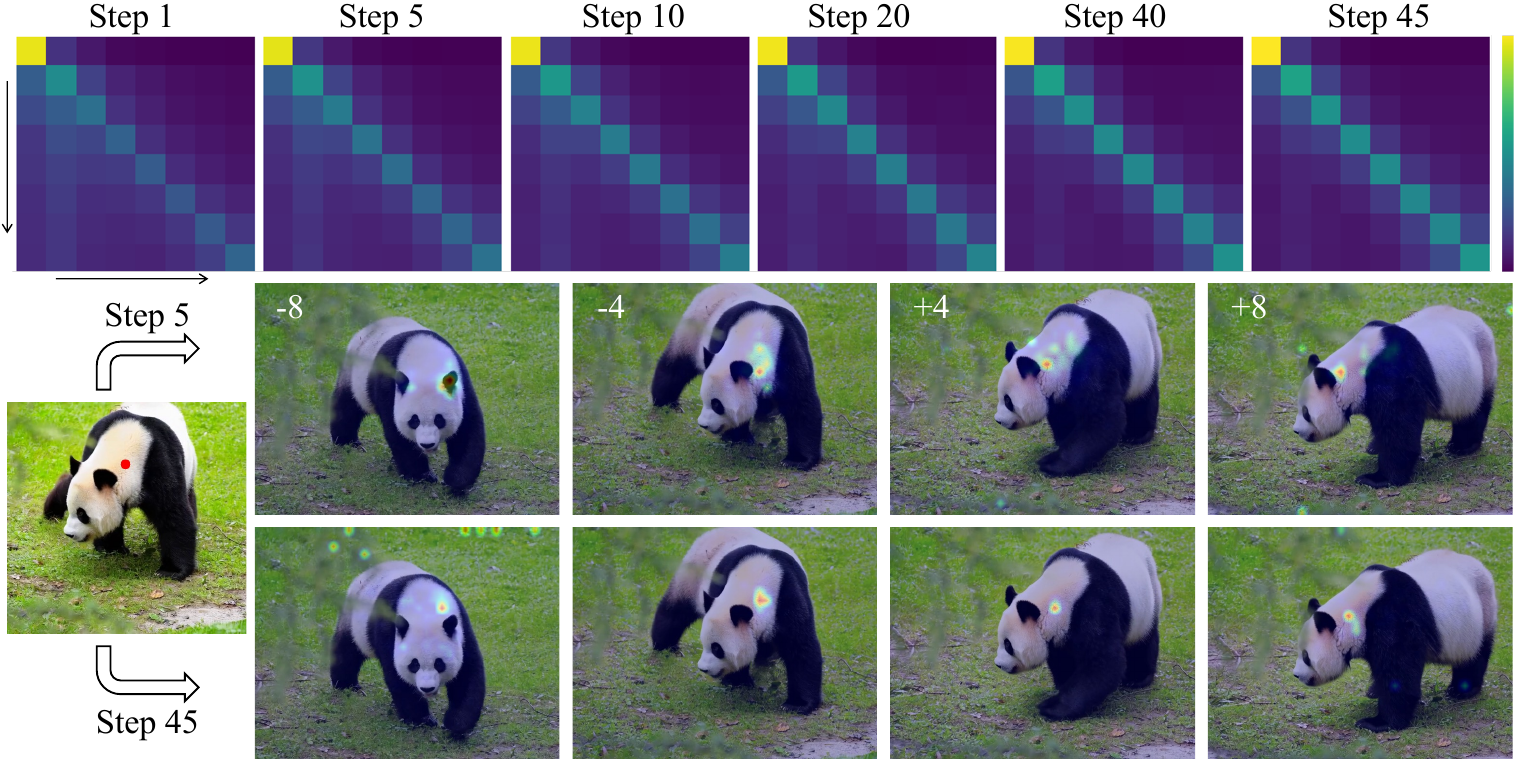}
\caption{
Visualization of cross-frame temporal (upper) and spatial (lower) attentions in different timesteps. }
\label{fig:attention_vis}
\vspace{-2ex}  
\end{wrapfigure}

Recent advances in video generation \cite{kong2024hunyuanvideo, wan2025wan} have seen a shift in architectural design from UNet-based models \cite{svd} to Transformer-based DiT frameworks \cite{peebles2023scalable}.
For finer-grained analysis of attention patterns, we decompose the cross-frame attention in the I2V Wan model into temporal and spatial components . To visualize temporal attention, we spatially average the attention values across each frame.
For spatial attention, we select a point on the query frame and record the distribution of attention intensities across other frames. As shown in Figure \ref{fig:attention_vis}, the model exhibits an auto-causal attention pattern, with lower attention values in the upper triangular region of the temporal attention matrices. Notably, the second frame (immediately following the reference frame) consistently receives high attention throughout the entire denoising process. This suggests that the reference frame can influence the generation of all subsequent frames indirectly, by strongly guiding the denoising of the second frame.

During late stages of denoising, the model concentrates on refining local details, such as textures, resulting in more accurate pixel-level correspondences compared to earlier stages that primarily require global information for layout establishment. This is reflected in the sharper and more focused high-attention regions. However, when attending to frames that are temporally distant from the query frame, the pixel-level attention tends to diffuse into incorrect areas, likely due to increased motion (camera or objects) between distant frames relative to adjacent ones. Note that accurate pixel correspondences are crucial for preserving the consistency of local stylized textures or strokes in reference-based video stylization. Therefore, an external control strategy is required to regularize the attention area for better appearance preservation. 
\vspace{-2ex}

\section{Method}
\label{sec:method}
\begin{figure}[!t]
    \vspace{-1ex}
    \includegraphics[width=\linewidth]{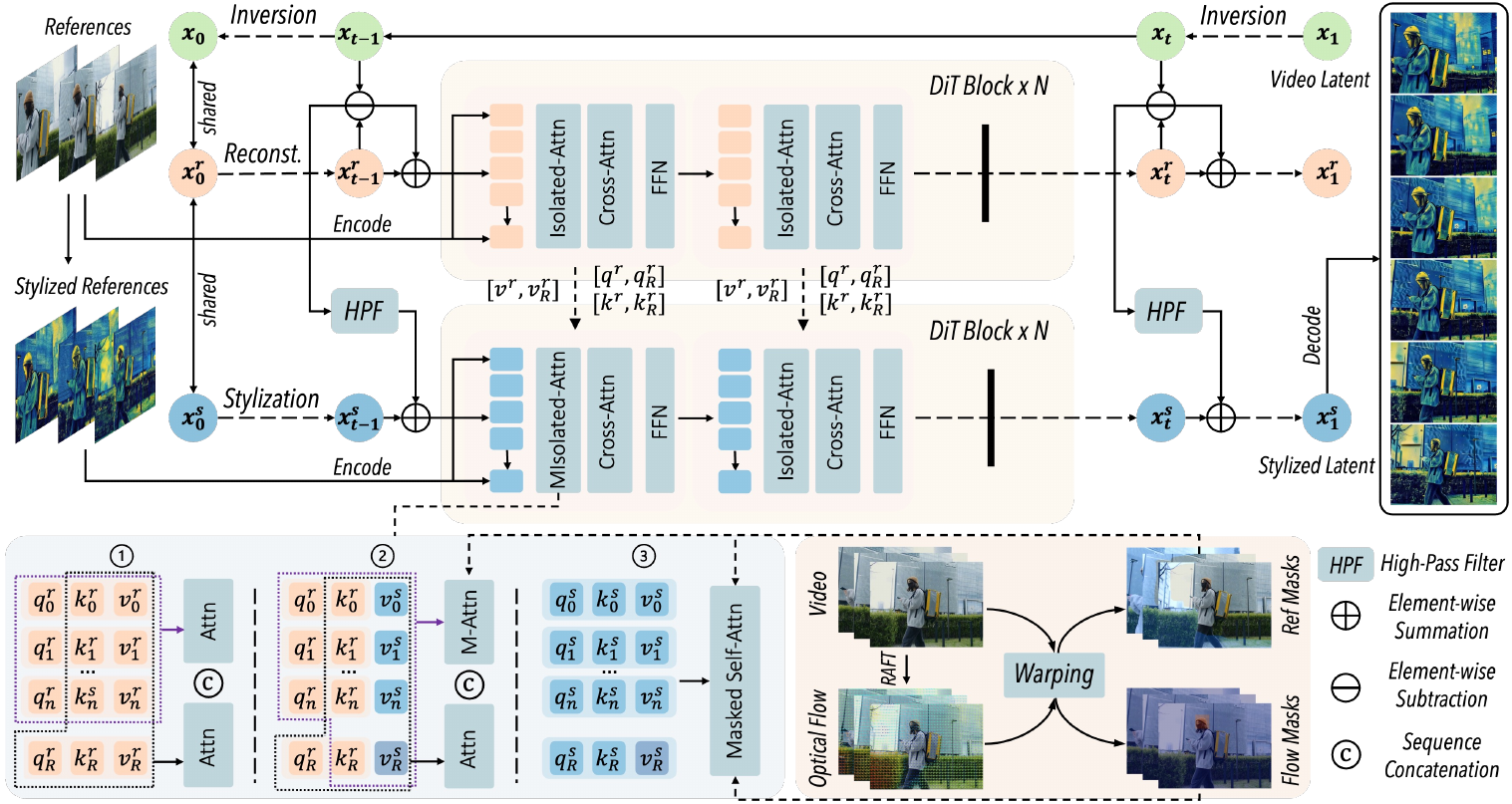}
    \vspace{-4ex}
    \caption{
    Overview of the FreeViS pipeline. \textbf{Isolated-Attn} indicates the \ding{172} mode, while \textbf{MIsolated-Attn} includes \ding{173} and \ding{174} attention modes. Optical flow, extracted using RAFT \cite{teed2020raft}, generates reference and flow masks for masked attention in attention modes \ding{173} and \ding{174}. 
    \vspace{-1ex}
}
    \label{fig:method}
    \vspace{-2ex}
\end{figure}

\vspace{-1ex}
The overall architecture of the proposed FreeViS pipeline is illustrated in Figure~\ref{fig:method}. Our framework builds upon a pretrained I2V diffusion model, which serves as the backbone. Given an arbitrary style image, stylized references are generated via an image style transfer model applied to several selected content video frames. We leverage inversion to recover the denoising trajectory and initial noise. The pipeline comprises two branches: reconstruction and stylization. Each branch is conditioned on the corresponding selected references.
The reconstruction branch provides query, key, and value matrices to the stylization branch in every DiT block. Parameters specific to the reconstruction and stylization branches are denoted by superscripts $r$ and 
$s$, respectively. The subscript $R$ denotes parameters associated with additional references beyond the first frame. After each denoising step, the high-frequency components of the difference between the target latent $\mathbf{x}_t$ and reconstruction latent $\mathbf{x}^r_t$ are added to the stylization latent $\mathbf{x}^s_t$. The value matrices $\mathbf{v}^s_R$ projected from additional stylized references are injected with dynamic clues, as formalized in Equation~\ref{eq:dynamic}.
\vspace{-2ex}

\subsection{Indirect High-frequency Compensation}
We adopt PnP Inversion \cite{ju2024pnp} to provide additional guidance throughout the denoising process. Specifically, the full denoising trajectory $[\mathbf{x}_t, \mathbf{x}_{t-1}, \ldots, \mathbf{x}_0]$ is cached during inversion and treated as the target latents. In the original PnP framework, the difference between the target latent $\mathbf{x}_t\in\mathbb{R}^{s\times c\times h \times w}$, where $s$ is the number of encoded latent maps, and the current reconstruction latent $\mathbf{x}_t^{r}\in\mathbb{R}^{s\times c\times h \times w}$ is computed and applied to both the reconstruction and editing branches, enabling near-exact reconstruction. However, we observe that this strong correction directly drives all the information of the stylized latent toward the original target, resulting in undesirable color distribution shifts from the stylized video back toward the original content.

As analyzed in Section~\ref{sec:frequency}, low-frequency (LF) latents primarily control appearance and color, while high-frequency (HF) components encode layout and motion. This motivates our Indirect High-frequency Compensation (IHC) strategy, which injects only HF differences into the stylized latents, preserving the stylized appearance while correcting structural inconsistencies. In the reconstruction branch, full compensation is applied to recover video content: $\mathbf{x}_{t}^{r} = \lambda \cdot (\mathbf{x}_{t} - \mathbf{x}_{t}^{r}) + \mathbf{x}_{t}^{r}$,
where $\lambda$ is a hyperparameter linearly decaying over timesteps. To further align the color distribution, we firstly apply AdaIN \cite{huang2017arbitrary} on both $\mathbf{x}_{t}$ and $\mathbf{x}_{t}^{r}$ before calculating their difference:
\begin{equation}
    \mathcal{T}(\mathbf{x}_t)=\sigma(\mathbf{x}_{t}^{s})\left(\frac{\mathbf{x}_t-\mu(\mathbf{x}_t)}{\sigma(\mathbf{x}_t)}\right)+\mu(\mathbf{x}_{t}^{s}), \qquad \mathcal{T}(\mathbf{x}_{t}^{r})=\sigma(\mathbf{x}_{t}^{s})\left(\frac{\mathbf{x}_{t}^{r}-\mu(\mathbf{x}_{t}^{r})}{\sigma(\mathbf{x}_{t}^{r})}\right)+\mu(\mathbf{x}_{t}^{s}),
\end{equation}
where $\mathbf{x}_{t}^{s}$ refers to the stylization latents in current timestep $t$, and $\mu(\cdot)$ and $\sigma(\cdot)$ denote channel-wise mean and standard
deviation, respectively. 
We then compute the Fast Fourier Transform (FFT) of the reconstruction difference along spatial dimensions to obtain the frequency representation. A low-pass filter $H_{LP}$ is applied to isolate HF components, which are subsequently transformed back via inverse FFT (iFFT) and added to the stylized latent:
\begin{equation}
\mathbf{x}_{t}^{s} = \lambda \cdot \mathcal{F}^{-1} \left( \mathcal{F}\left(\mathcal{T}(\mathbf{x}_{t}) - \mathcal{T}(\mathbf{x}_{t}^{r})\right) \cdot (1 - H_{LP}) \right) + \mathbf{x}_{t}^{s},
\end{equation}
where $\mathcal{F}(\cdot)$ and $\mathcal{F}^{-1}(\cdot)$ denote FFT and iFFT. Empirically, IHC successfully preserves the stylized color and texture while enhancing spatial
consistency and motion fidelity, particularly in scenes with significant camera movement or when later frames differ substantially from the first frame. 

\vspace{-1ex}
\subsection{Additional Inconsistent References}
\label{sec:references}
Previous I2V-based video editing methods \cite{ku2024anyv2v, ouyang2024i2vedit} typically rely on a single stylized first frame to guide the stylization of the entire video. While effective when content changes are minimal, this approach struggles to propagate style features to regions that differ significantly from the first frame. An intuitive solution is to incorporate multiple stylized references. However, existing advanced I2V models support only a single reference input. 
Naively concatenating additional reference tokens with reconstruction or stylization noise tokens prior to self-attention often results in significant \textbf{flickering} and \textbf{stuttering} in the generated videos. 

To enable multi-reference inputs in the reconstruction branch, we propose the Isolated-Attn strategy,
which isolates the influence of auxiliary references $\mathbf{x}^{r}_{R}$, thereby preventing interference with the original denoising schedule. Specifically, the reconstruction tokens $\mathbf{x}^r$ undergo standard self-attention, while the reference tokens $\mathbf{x}^{r}_{R}$ attend to both reconstruction and reference keys and values. This design allows reference tokens to evolve in sync with the denoising process, mimicking full self-attention behavior. Denote the Query, Key, and Value matrices projected from $\mathbf{x}^r$ and $\mathbf{x}^{r}_{R}$ as $\{\mathbf{Q}^r,\mathbf{K}^r,\mathbf{V}^r\}$ and $\{\mathbf{Q}^r_R,\mathbf{K}^r_R,\mathbf{V}^r_R\}$, respectively. Then, this mechanism can be defined as:
\begin{equation}
    \mathbf{Out}^{r}=\mathcal{A}(\mathbf{Q}^r,\mathbf{K}^r,\mathbf{V}^r)\oplus \mathcal{A}(\mathbf{Q}_{R}^{r},\mathbf{K}^r\oplus \mathbf{K}_{R}^{r},\mathbf{V}^r\oplus \mathbf{V}_{R}^{r}),
\end{equation}
where $\mathcal{A}(\cdot)$ refers to attention operation, and $\oplus$ denotes sequence-wise token concatenation. 

Differently, in the stylization branch, full information exchange among all tokens is essential for comprehensive style transfer. However, since the additional stylized references $\mathbf{x}^{s}_{R}$ are independently encoded, the value matrices $\mathbf{V}^{s}_{R}$ projected from $\mathbf{x}^{s}_{R}$ lack dynamic information, which causes the \textbf{stuttering} issue in adjacent frames. We observe that appearance information can be isolated and exchanged between videos with shared dynamics (see details in Appendix \ref{app:decomposition}).
Since the dynamic information is shared between stylization values $\mathbf{V}^s$ and reconstruction values $\mathbf{V}^r$, we decouple the dynamic residual by computing their differences relative to their respective references (only contain appearance), and inject only the dynamic component into $\mathbf{V}^{s}_{R}$. This is implemented as:
\begin{equation}
\label{eq:dynamic}
\mathbf{V}_{R}^{s} = \mathbf{V}_{R}^{s} + \xi \cdot \left( \mathbf{V}^{s}[i_R] - \mathbf{V}_{R}^{s} \right) + (1 - \xi) \cdot \left( \mathbf{V}^{r}[i_R] - \mathbf{V}_{R}^{r} \right),
\end{equation}
where $i_R$ denotes the positional indices of the references, and $\xi$ increases linearly from 0 to 1 over the timesteps. Early in the denoising process, the model relies more on dynamics from the reconstruction branch; toward the end, $\mathbf{V}_{R}^{s}$ converges to $\mathbf{V}^{s}[i_R]$ to ensure consistent final outputs. 

Since inconsistencies naturally exist across stylized references, the same region may exhibit varying appearances, resulting in time-variant stylization artifacts, such as \textbf{flickering}. To address this issue, we leverage optical flow extracted from the content video to identify regions already covered by earlier references and mask these regions in attention, thereby resolving appearance conflicts. We employ RAFT \cite{teed2020raft} to compute optical flow and trace pixel correspondences from the first reference frame to subsequent reference frames. The reference mask $M_{Ref}$ is constructed such that a position is marked as \texttt{False} if the corresponding pixel is reachable from previous references(see Appendix \ref{app:ref_mask}). Masked attention $\mathcal{A}_{Masked}(\cdot)$ is implemented to incorporate $M_{Ref}$:
\begin{equation}
\label{eq:out1}
    \mathbf{Out}_{1}^{s}=\mathcal{A}_{Masked}(\mathbf{Q}^s,\mathbf{K}^s\oplus \mathbf{K}_{R}^{s},\mathbf{V}^s\oplus \mathbf{V}_{R}^{s}, M_{Ref})\oplus \mathcal{A}(\mathbf{Q}^s_{R},\mathbf{K}^s\oplus \mathbf{K}_{R}^{s},\mathbf{V}^s\oplus \mathbf{V}_{R}^{s}),
\end{equation}
where $\{\mathbf{Q}^s,\mathbf{K}^s\}$ and $\{\mathbf{Q}^s_R,\mathbf{K}^s_R\}$ refer to the Query and Key matrices projected by $\mathbf{x}^s$ and $\mathbf{x}^s_R$. Inspired by prior UNet-based approaches that forward convolution and attention features from reconstruction to editing branches, we leverage QK-Sharing of the reconstruction queries and keys to enable more flexible attention operations in the stylization branch.
These queries and keys define spatial-temporal correspondences across the content video, which are critical for effective style propagation and temporal consistency. Thus, we replace the query and value in Equation \ref{eq:out1}:
\begin{equation}
    \mathbf{Out}_{2}^{s}=\mathcal{A}_{Masked}(\mathbf{Q}^r,\mathbf{K}^r\oplus \mathbf{K}_{R}^{r},\mathbf{V}^s\oplus \mathbf{V}_{R}^{s}, M_{Ref})\oplus \mathcal{A}(\mathbf{Q}^r_{R},\mathbf{K}^r\oplus \mathbf{K}_{R}^{r},\mathbf{V}^s\oplus \mathbf{V}_{R}^{s})
\end{equation}
 Due to the auto-causal nature of the base model, the style of the generated video is primarily determined by the first reference frame, while subsequent references act as supplementary sources.

\vspace{-1ex}
\subsection{Explicit Optical Flow Guidance}
When the camera or object motion is significant, we observe disappearing or time-varying style textures in plain areas. We attribute this issue to inaccurate attention maps between temporally distant frames, especially in regions with little salient visual features, as discussed in Section~\ref{sec:cross_frame}. To address this, we introduce Explicit Optical-flow Guidance (EOG), which provides pixel-level correspondence via optical flow. We first compute forward and backward optical flows and trace every pixel across frames. If a pixel $p_{i,j}^{s}$ in frame $s$ maps to location $p_{m,n}^{t}$ in frame $t$, the flow mask $M_{Flow} \in \{0,1\}^{T \times h \times w \times T \times h \times w}$ is set to \texttt{True} at index $(s,i,j,t,m,n)$ (see Appendix~\ref{app:flow_mask}). We further apply dilation to compensate for estimation errors of the optical flow.

Using this mask, we perform masked attention on stylization and corresponding reference tokens:
\begin{equation}
    \mathbf{Out}_{3}^{s}=\mathcal{A}_{Masked}(\mathbf{Q}^s\oplus \mathbf{Q}_{R}^{s},\mathbf{K}^s\oplus \mathbf{K}_{R}^{s},\mathbf{V}^s\oplus \mathbf{V}_{R}^{s}, M_{Flow}\land M_{Ref}),
\end{equation}
where $\land$ denotes a logical AND, ensuring that only consistent regions across flow and reference masks contribute. This design constrains the attention area of the source pixel on every frame, explicitly controlling the appearance to follow the video dynamics. Empirically, we find that using queries and keys from stylization tokens outperforms those from the reconstruction branch. Finally, the outputs from all three attention modes are aggregated before cross-attention as:
\begin{equation}
    \mathbf{Out}^{s}=(1-\beta - \gamma)\cdot \mathbf{Out}_{1}^{s} + \beta \cdot \mathbf{Out}_{2}^{s} + \gamma \cdot \mathbf{Out}_{3}^{s},
\end{equation}
where $\beta$ and $\gamma$ are two hyperparameters, controlling the impact of the inconsistent references and the strength of the explicit appearance guidance. $\gamma$ is set to a non-zero value only in the final stage of the denoising process, when the model primarily focuses on local texture refinement. 

\textbf{Cross-Attention.} The base model further encodes the reference image using CLIP \cite{radford2021learning} to provide high-level semantic guidance via cross-attention. To incorporate multiple references, we concatenate the CLIP features from all reference frames and apply QK-Sharing in the cross-attention layer, enhancing the injection of language-aligned information. As CLIP features contain limited spatial structure, reference and flow masks are not applied.

\textbf{Reference Arrangement.} In principle, all the frames can be stylized and used as references in FreeViS for comprehensive style coverage. Nevertheless, increasing the number of references significantly raises computational and memory costs. Considering that most advanced I2V models are limited to short videos (typically around 81 frames), we select the first, middle, and last frames as references, an empirically effective choice for achieving high-quality stylization in most cases. Each reference token is assigned the same positional embedding as its corresponding frame latent to guarantee correct spatial and temporal propagation of stylized features.

\vspace{-2ex}
\section{Experiments}
\newlength{\ts}\setlength{\ts}{5mm}
\newcommand{\tts}{\hspace{\ts}}
\begin{table}[!t]
\vspace{-1ex}
\centering
\small
\begin{tabular}{l*{4}{@{\tts}c}@{\tts}l*{2}{@{\tts}c}@{\tts}c}
\toprule
\multirow{2}{*}{\textbf{Method}} 
  & \multicolumn{4}{c}{\textbf{Stylization Quality}} 
  & \multicolumn{3}{c}{\textbf{Video Consistency}} 
  & \multirow{2}{*}{\textbf{HP} $\uparrow$} \\
\cmidrule(lr){2-5} \cmidrule(lr){6-8}
  & \textbf{CSD Score} $\uparrow$  
  & \textbf{ArtFID} $\downarrow$ 
  & \textbf{FID} $\downarrow$ 
  & \textbf{LPIPS} $\downarrow$ 
  & \textbf{SC} $\uparrow$
  & \textbf{MS} $\uparrow$
  & \textbf{FC} $\downarrow$ 
  & \\
\midrule
\rowcolor{blue!8}
Reference  & 0.508 & 31.62 & 20.28 & 0.486    & 0.918 & 0.986 & 0.000  & - \\
TokenFlow  & 0.111 & 37.87 & 27.94 & 0.309    & \textbf{0.915} & 0.976 & 1.092  & 2.179 \\
VACE       & 0.138 & 35.53 & 27.77 & 0.240    & 0.910 & \textbf{0.984} & \textbf{0.554}  & 2.895 \\
\rowcolor{gray!10}
I2VEdit    & 0.331 & 38.72 & 22.53 & 0.653    & 0.738 & 0.975 & 2.074  & 2.538 \\
\rowcolor{gray!10}
AnyV2V     & 0.267 & 35.84 & 23.52 & 0.471    & 0.753 & 0.961 & 1.715 & 2.443 \\
\rowcolor{gray!10}
AnyV2V*    & 0.270 & 34.81 & 27.59 & \textbf{0.218}    & 0.675 & 0.983 & 1.103 & 3.372 \\
\rowcolor{green!8}
Ours       & \textbf{0.448} & \textbf{33.46} & \textbf{21.62}     & 0.479    & 0.898 & 0.978 & 0.641     & \textbf{4.113} \\
\bottomrule
\end{tabular}
\vspace{-1ex}
\caption{
Quantitative results for video stylization. "SC", "MS", "FC", and "HP" denote Style Consistency, Motion Smoothness, Flow Consistency, and Human Preference, respectively. We use the stylized references and original video to anchor the quality of stylization and video consistency.
} 
\label{tab:1}
\vspace{-3ex}
\end{table}

This section is divided into two major parts: Video Style Transfer and Stylized T2V Generation. All the stylized references for FreeViS and other reference-based video editing models are provided by InstantStyle-plus \cite{wang2024instantstyle_plus}. We deliver the results of FreeViS combined with various image stylization methods and different video diffusion base models in Appendix \ref{app:other_image_stylization} and \ref{app:other_base_model}.

\vspace{-2ex}
\subsection{Video Style Transfer}
\textbf{Dataset \& Metrics.}
We curated 200 online videos covering human activities, natural and urban scenes, animals, and transportation. Style images were sourced from WikiArt \cite{tan2018improved} and web collections. For evaluation, we use the CSD Score \cite{somepalli2024measuring} to assess frame–style similarity, and ArtFID \cite{wright2022artfid} to jointly evaluate stylization quality and content preservation via FID and LPIPS. Motion smoothness is measured with VBench \cite{huang2024vbench}, and optical flow consistency between the original and stylized videos with End-Point Error (EPE) \cite{barron1994performance}. Finally, we introduce Style Consistency, which quantifies the similarity between the first frame and subsequent frames based on style features extracted by \cite{somepalli2024measuring}. We conduct a human preference evaluation to assess both the visual plausibility and the stylization quality of the generated videos. Each examiner is asked to assign a score from 1 to 5 for both aspects, and the final rating is obtained by averaging the two scores.

\begin{figure}[!t]
    \vspace{-1ex}
    \includegraphics[width=\linewidth]{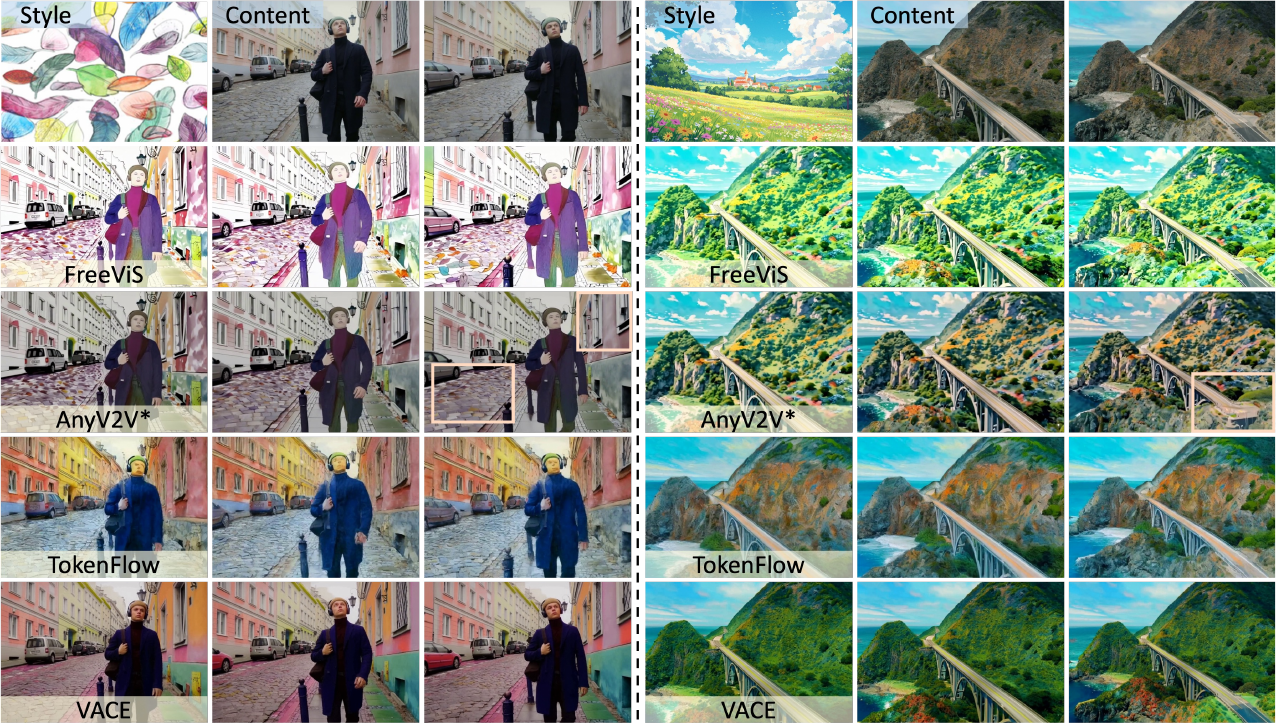}
    \vspace{-4ex}
    \caption{
    Qualitative comparison of FreeViS with other video editing methods on video stylization. The areas inside the bounding boxes show missing style textures and incorrect reconstruction.
}
    \label{fig:exp1}
    \vspace{-1ex}
\end{figure}

\begin{figure}[!t]
    \vspace{-1ex}
    \includegraphics[width=\linewidth]{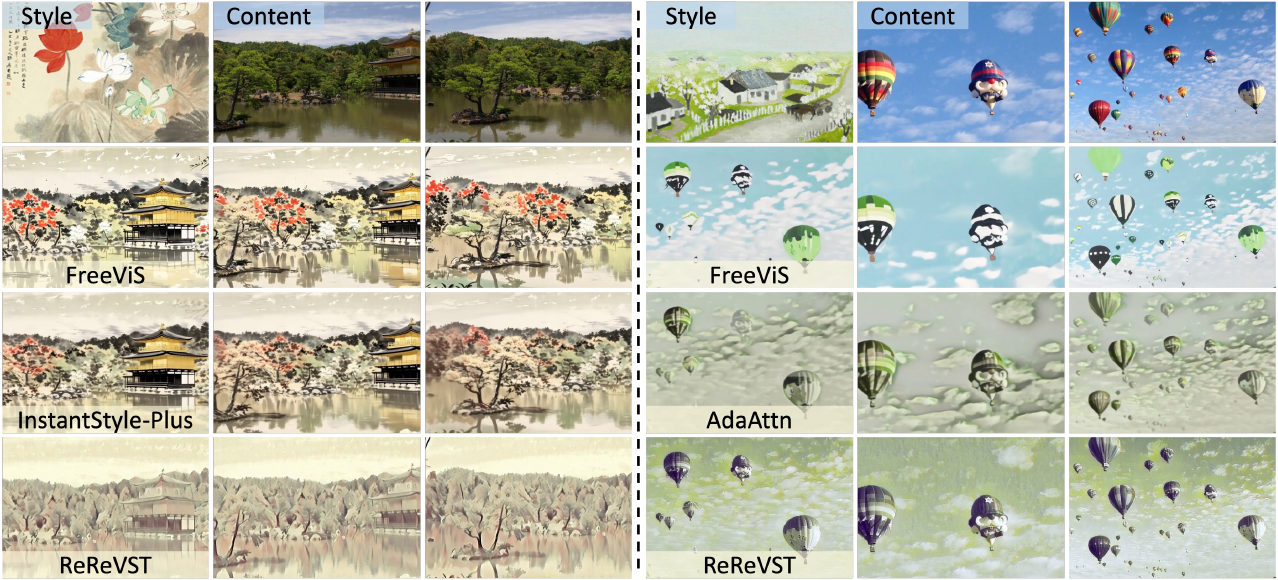}
    \vspace{-4ex}
    \caption{
    Qualitative comparison of FreeViS with previous video and image stylization methods. The flickering issue of the image stylization methods can be observed in the supplemented video.
}
    \label{fig:exp2}
    \vspace{-4ex}
\end{figure}

\textbf{Baselines.}
We compare against various baselines: reference-based video editing (AnyV2V \cite{ku2024anyv2v}, I2VEdit \cite{ouyang2024i2vedit}), text-driven editing (TokenFlow \cite{geyer2024tokenflow}), unified video editing (VACE \cite{jiang2025vace}), classic video style transfer (ReReVST \cite{wang2020consistent}, ViSt3D \cite{pande2023vist3d}), and image style transfer with temporal smoothing \cite{duan2023fastblend} (AdaAttn \cite{liu2021adaattn}, InstantStyle-plus \cite{wang2024instantstyle_plus}). 
Since the code for V2V implementation of StyleMaster \cite{ye2025stylemaster} has not been released, we omit its results currently. We re-implement AnyV2V, named by AnyV2V*, using the same base model as FreeViS for a fair comparison. Note that among all the models, only AnyV2V and TokenFlow are training-free.

\textbf{Results.}
Qualitative comparisons are presented in Figures \ref{fig:exp1} and \ref{fig:exp2}. VACE demonstrates strong video consistency (Table \ref{tab:1}) but is limited to color modification, struggling to capture and transfer complex style textures. Similarly, TokenFlow faces challenges in style representation, primarily due to the limitations of text-based descriptions, which parallel the issues observed with VACE. The original AnyV2V and I2VEdit models fail to reconstruct the video when there are substantial content changes between the first and last frames (see Appendix \ref{app:more_v2v}), likely due to the weak priors of their pre-trained base models. While AnyV2V* significantly mitigates this issue, it still suffers from propagation errors: early-frame styles are not consistently transferred to later frames, resulting in the lowest style consistency score. In contrast, FreeViS effectively preserves stylistic features across the entire sequence, achieving the highest stylization quality and near-optimal temporal consistency.

Classical video stylization methods, which typically rely on non-diffusion architectures, offer strong temporal consistency but suffer from limited stylization quality. In contrast, image-based stylization methods, while capable of richer stylization, exhibit significant frame-wise inconsistencies. This leads to over-flattened style textures and persistent flickering artifacts, even after temporal smoothing. FreeViS overcomes these limitations by leveraging a pretrained video diffusion framework that effectively balances high-quality stylization with robust temporal coherence.

\begin{figure}[!t]
    \vspace{-1ex}
    \includegraphics[width=\linewidth]{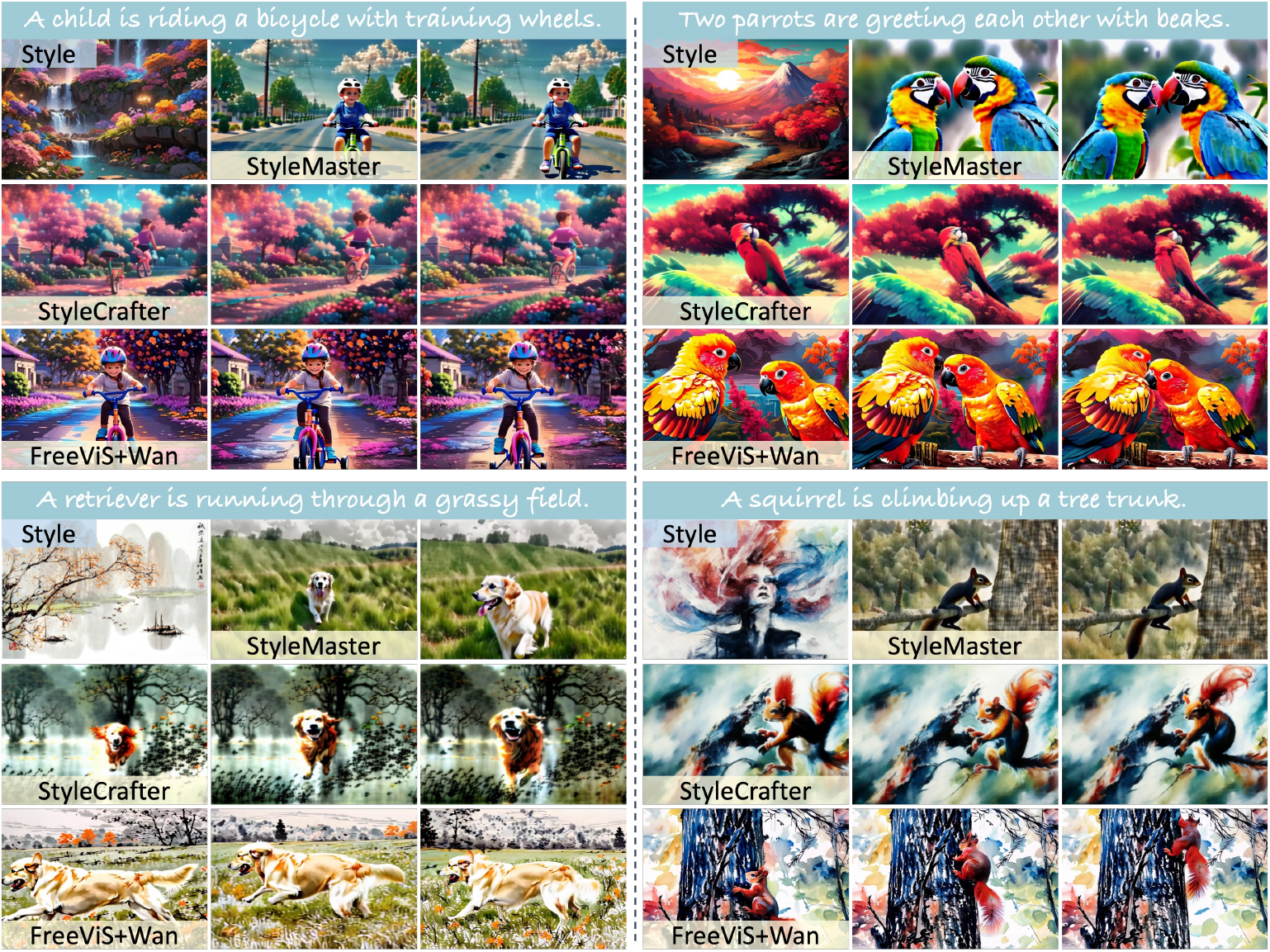}
    \vspace{-4ex}
    \caption{
    Qualitative comparison of FreeViS with previous methods for stylized video generation. The difference can be easily observed in the supplemented video.
}
    \label{fig:exp3}
    \vspace{-1ex}
\end{figure}
\vspace{-1ex}
\begin{table}[!t]
\vspace{-1ex}
\centering
\small
\begin{tabular}{l*{3}{@{\tts}c}@{\tts}l*{3}{@{\tts}c}@{\tts}c}
\toprule
\multirow{2}{*}{\textbf{Method}} 
  & \multicolumn{2}{c}{\textbf{Stylization Quality}} 
  & \multicolumn{5}{c}{\textbf{Video Generation Quality}} 
  & \multirow{2}{*}{\textbf{HP} $\uparrow$} \\
\cmidrule(lr){2-3} \cmidrule(lr){4-8}
  & \textbf{CSD Score} $\uparrow$
  & \textbf{FID} $\downarrow$ 
  & \textbf{CLIP-Text} $\uparrow$ 
  & \textbf{DQ} $\uparrow$
  & \textbf{MS} $\uparrow$
  & \textbf{BC} $\uparrow$
  & \textbf{IQ} $\uparrow$
  & \\
\midrule
\rowcolor{gray!10}
StyleCrafter    & \textbf{0.515} & \textbf{22.62} & 0.211 & 0.368    & 0.965 & \textbf{0.951}  & 0.578   & 2.83 \\
\rowcolor{gray!10}
StyleMaster     & 0.221 & 26.04 & 0.243 & 0.123    & \textbf{0.985} & 0.945  & 0.667   & 2.55 \\
\rowcolor{green!8}
Ours+Wan       & 0.437 & 24.63 & \textbf{0.264}        & \textbf{0.509} & 0.980 & 0.941  & \textbf{0.691}   & \textbf{3.97} \\
\bottomrule
\end{tabular}
\vspace{-1ex}
\caption{
Quantitative results for stylized video generation. "DQ", "MS", "BC", "IQ", and "HP" denote Dynamic Quality, Motion Smoothness, Background (CLIP) Consistency, Imaging Quality, and Human Preference, respectively. A smaller DQ can naturally lead to better MS and BC.
} 
\label{tab:2}
\vspace{-4ex}
\end{table}

\vspace{-1ex}
\subsection{Stylized T2V Generation}
\textbf{Dataset \& Metrics \& Baselines.}
We utilize Qwen3 \cite{yang2025qwen3} to generate 200 text prompts describing the content of target videos, covering a broad range of commonly encountered real-world scenarios. The style images used are identical to those described in the previous section. As no ground-truth videos are available, we adopt four evaluation metrics from VBench \cite{huang2024vbench}: Dynamic Quality, Motion Smoothness, Background Consistency, and Imaging Quality, to assess the visual quality of the generated videos. Additionally, CLIP-Text similarity is employed to evaluate video-prompt alignment. For assessing style consistency and fidelity, we report the CSD Score and FID, which measure the similarity between video frames and style images. Human preference is also employed to assess the prompt alignment, stylization quality, and visual plausibility. We compare our approach against two state-of-the-art baselines \cite{liu2024stylecrafter,ye2025stylemaster}: StyleCrafter and StyleMaster. For stylized T2V generation, we first generate base videos using Wan2.1 \cite{wan2025wan}, followed by applying FreeViS for video stylization.

\textbf{Results.}
The qualitative and quantitative results for stylized video generation are presented in Figure \ref{fig:exp3} and Table \ref{tab:2}. StyleCrafter effectively transfers style into the generated videos; however, its outputs often lack dynamic realism and occasionally fail to align with the input prompts. In contrast, StyleMaster better adheres to the textual prompts but exhibits limited stylization capability. The combination of FreeViS and the Wan model achieves the best trade-off between stylization fidelity and content alignment, producing videos that are both visually compelling and semantically accurate, as reflected by the highest human preference scores. Note that the stylization performance of FreeViS is inherently upper-bounded by the underlying image style transfer method it employs.

\begin{figure}[!t]
    \vspace{-1ex}
    \includegraphics[width=\linewidth]{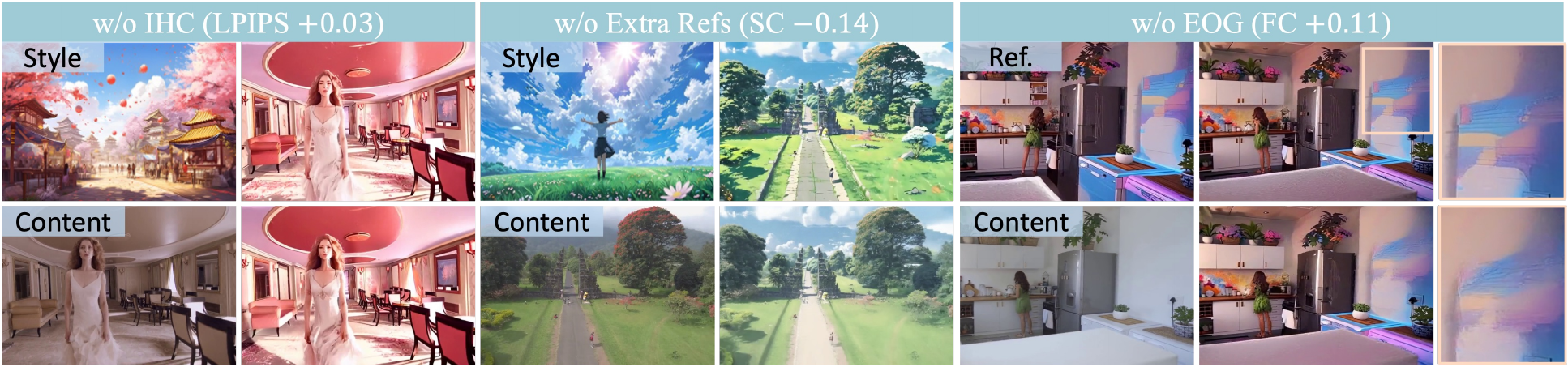}
    \vspace{-4ex}
    \caption{
    Ablation study of FreeViS. The upper images are generated from the full model.
}
    \label{fig:ablation}
    \vspace{-3ex}
\end{figure}

\vspace{-2ex}
\subsection{Ablation Study}
\vspace{-1ex}
FreeViS primarily comprises three key components: IHC for layout reconstruction, Extra References for enforcing style consistency, and EOG for texture preservation in plain areas. The effectiveness of each component is illustrated in Figure \ref{fig:ablation}. Without IHC, the model fails to accurately reconstruct detailed scene layouts, leading to structural artifacts. Incorporating additional reference frames helps maintain consistent style features throughout the video, significantly enhancing visual coherence. Furthermore, the EOG module mitigates the loss of style textures in visually homogeneous regions of the original video, thereby improving flow consistency and overall stylization quality.

\vspace{-2ex}
\section{Related Work}
\textbf{Video Diffusion.}
Diffusion-based video generation is advancing rapidly, with significant progress in quality~\cite{agarwal2025cosmos, yang2025cogvideox, svd}, efficiency~\cite{xi2025sparse, zhang2025flashvideo, kahatapitiya2024adaptive}, controllability~\cite{zhang2025enabling, bai2025recammaster, ren2025gen3c}, and extrapolation~\cite{zhao2025riflex, dalal2025one, mao2024story}. Unlike early UNet-based architectures~\cite{svd, guo2024animatediff, mei2023}, which employ separate layers to model spatial and temporal dependencies, recent advances~\cite{kong2024hunyuanvideo, yang2025cogvideox, wan2025wan} adopt the full self-attention layers of DiT~\cite{peebles2023scalable} to jointly capture pixel-level coherence.

\textbf{Video Editing.}
Leveraging the strong priors of video diffusion models, recent editing methods have significantly outperformed earlier approaches based on optical flow \cite{xu2019deep}, GANs \cite{tzaban2022stitch,fruhstuck2023vive3d}, and image diffusion \cite{ceylan2023pix2video,geyer2024tokenflow,feng2024ccedit}. Unlike image editing, which mainly enforces spatial coherence \cite{Brooks_2023_CVPR,brooks2023instructpix2pix}, video editing additionally requires preserving temporal consistency \cite{yang2025videograin}. Prior work has addressed diverse tasks such as inpainting \cite{bian2025videopainter,guo2025keyframe}, content swapping \cite{liu2024video,tu2025videoanydoor,zhang2025fastvideoedit}, and interactive control \cite{deng2024dragvideo,ma2025magicstick}. VACE \cite{jiang2025vace} proposes a unified framework for video generation and editing with a mask-guided condition unit, while AnyV2V \cite{ku2024anyv2v} and I2VEdit \cite{ouyang2024i2vedit} leverage the first edited frame as guidance for pre-trained I2V models. 

\textbf{Style Transfer.}
Neural style transfer~\cite{gatys2016image} has progressed with deep architectures~\cite{huang2017arbitrary, liu2021adaattn, deng2022stytr2} and generative models~\cite{azadi2018multi, zhang2023inversion, wang2023stylediffusion}. Diffusion models further improve semantic alignment between style and content~\cite{wang2024instantstyle_plus, xing2024csgo}, inspiring training-free methods~\cite{wang2024instantstyle, rout2025rb, chung2024style, xu2025stylessp} that surpass traditional techniques. By contrast, video stylization remains underexplored. ViSt3D~\cite{pande2023vist3d} employs 3D CNNs for temporal consistency, StyleMaster~\cite{ye2025stylemaster} finetunes a video diffusion model with CLIP-guided cross-attention, and StyleCrafter~\cite{liu2024stylecrafter} enables style-conditioned text-to-video generation. Nonetheless, these approaches still lag behind image style transfer in quality, with temporal consistency being a key challenge.

\vspace{-2ex}
\section{Conclusion}
\vspace{-1ex}
In this paper, we introduce a training-free video stylization framework, FreeViS, which successfully addresses the propagation errors inherent in prior works. FreeViS leverages IHC to constrain the layout and motion. To enable additional reference inputs without requiring extra fine-tuning, we propose isolated attention with masking and dynamics injection. Furthermore, we propose the EOG strategy to preserve style textures in plain areas. Extensive experiments are conducted to validate the effectiveness of our design and the superior performance of FreeViS over other methods.

\bibliography{iclr2026_conference}

\begin{thebibliography}{76}
\providecommand{\natexlab}[1]{#1}
\providecommand{\url}[1]{\texttt{#1}}
\expandafter\ifx\csname urlstyle\endcsname\relax
  \providecommand{\doi}[1]{doi: #1}\else
  \providecommand{\doi}{doi: \begingroup \urlstyle{rm}\Url}\fi

\bibitem[Agarwal et~al.(2025)Agarwal, Ali, Bala, Balaji, Barker, Cai, Chattopadhyay, Chen, Cui, Ding, et~al.]{agarwal2025cosmos}
Niket Agarwal, Arslan Ali, Maciej Bala, Yogesh Balaji, Erik Barker, Tiffany Cai, Prithvijit Chattopadhyay, Yongxin Chen, Yin Cui, Yifan Ding, et~al.
\newblock Cosmos world foundation model platform for physical ai.
\newblock \emph{arXiv preprint arXiv:2501.03575}, 2025.

\bibitem[Azadi et~al.(2018)Azadi, Fisher, Kim, Wang, Shechtman, and Darrell]{azadi2018multi}
Samaneh Azadi, Matthew Fisher, Vladimir~G Kim, Zhaowen Wang, Eli Shechtman, and Trevor Darrell.
\newblock Multi-content gan for few-shot font style transfer.
\newblock In \emph{Proceedings of the IEEE conference on computer vision and pattern recognition}, pp.\  7564--7573, 2018.

\bibitem[Bai et~al.(2025{\natexlab{a}})Bai, Xia, Fu, Wang, Mu, Cao, Liu, Hu, Bai, Wan, et~al.]{bai2025recammaster}
Jianhong Bai, Menghan Xia, Xiao Fu, Xintao Wang, Lianrui Mu, Jinwen Cao, Zuozhu Liu, Haoji Hu, Xiang Bai, Pengfei Wan, et~al.
\newblock Recammaster: Camera-controlled generative rendering from a single video.
\newblock \emph{arXiv preprint arXiv:2503.11647}, 2025{\natexlab{a}}.

\bibitem[Bai et~al.(2025{\natexlab{b}})Bai, Chen, Liu, Wang, Ge, Song, Dang, Wang, Wang, Tang, Zhong, Zhu, Yang, Li, Wan, Wang, Ding, Fu, Xu, Ye, Zhang, Xie, Cheng, Zhang, Yang, Xu, and Lin]{bai2025qwen25vltechnicalreport}
Shuai Bai, Keqin Chen, Xuejing Liu, Jialin Wang, Wenbin Ge, Sibo Song, Kai Dang, Peng Wang, Shijie Wang, Jun Tang, Humen Zhong, Yuanzhi Zhu, Mingkun Yang, Zhaohai Li, Jianqiang Wan, Pengfei Wang, Wei Ding, Zheren Fu, Yiheng Xu, Jiabo Ye, Xi~Zhang, Tianbao Xie, Zesen Cheng, Hang Zhang, Zhibo Yang, Haiyang Xu, and Junyang Lin.
\newblock Qwen2.5-vl technical report, 2025{\natexlab{b}}.
\newblock URL \url{https://arxiv.org/abs/2502.13923}.

\bibitem[Barron et~al.(1994)Barron, Fleet, and Beauchemin]{barron1994performance}
John~L Barron, David~J Fleet, and Steven~S Beauchemin.
\newblock Performance of optical flow techniques.
\newblock \emph{International journal of computer vision}, 12\penalty0 (1):\penalty0 43--77, 1994.

\bibitem[Bian et~al.(2025)Bian, Zhang, Ju, Cao, Xie, Shan, and Xu]{bian2025videopainter}
Yuxuan Bian, Zhaoyang Zhang, Xuan Ju, Mingdeng Cao, Liangbin Xie, Ying Shan, and Qiang Xu.
\newblock Videopainter: Any-length video inpainting and editing with plug-and-play context control.
\newblock In \emph{Proceedings of the Special Interest Group on Computer Graphics and Interactive Techniques Conference Conference Papers}, pp.\  1--12, 2025.

\bibitem[Blattmann et~al.(2023)Blattmann, Dockhorn, Kulal, Mendelevitch, Kilian, Lorenz, Levi, English, Voleti, Letts, et~al.]{svd}
Andreas Blattmann, Tim Dockhorn, Sumith Kulal, Daniel Mendelevitch, Maciej Kilian, Dominik Lorenz, Yam Levi, Zion English, Vikram Voleti, Adam Letts, et~al.
\newblock Stable video diffusion: Scaling latent video diffusion models to large datasets.
\newblock \emph{arXiv preprint arXiv:2311.15127}, 2023.

\bibitem[Brooks et~al.(2023{\natexlab{a}})Brooks, Holynski, and Efros]{Brooks_2023_CVPR}
Tim Brooks, Aleksander Holynski, and Alexei~A. Efros.
\newblock Instructpix2pix: Learning to follow image editing instructions.
\newblock In \emph{Proceedings of the IEEE/CVF Conference on Computer Vision and Pattern Recognition (CVPR)}, pp.\  18392--18402, June 2023{\natexlab{a}}.

\bibitem[Brooks et~al.(2023{\natexlab{b}})Brooks, Holynski, and Efros]{brooks2023instructpix2pix}
Tim Brooks, Aleksander Holynski, and Alexei~A Efros.
\newblock Instructpix2pix: Learning to follow image editing instructions.
\newblock In \emph{Proceedings of the IEEE/CVF conference on computer vision and pattern recognition}, pp.\  18392--18402, 2023{\natexlab{b}}.

\bibitem[Cao et~al.(2023)Cao, Wang, Qi, Shan, Qie, and Zheng]{cao2023masactrl}
Mingdeng Cao, Xintao Wang, Zhongang Qi, Ying Shan, Xiaohu Qie, and Yinqiang Zheng.
\newblock Masactrl: Tuning-free mutual self-attention control for consistent image synthesis and editing.
\newblock In \emph{Proceedings of the IEEE/CVF international conference on computer vision}, pp.\  22560--22570, 2023.

\bibitem[Ceylan et~al.(2023)Ceylan, Huang, and Mitra]{ceylan2023pix2video}
Duygu Ceylan, Chun-Hao~P Huang, and Niloy~J Mitra.
\newblock Pix2video: Video editing using image diffusion.
\newblock In \emph{Proceedings of the IEEE/CVF International Conference on Computer Vision}, pp.\  23206--23217, 2023.

\bibitem[Chen et~al.(2025)Chen, Guo, Zhu, Zhang, Huang, Feng, and Kang]{chen2025video}
Sili Chen, Hengkai Guo, Shengnan Zhu, Feihu Zhang, Zilong Huang, Jiashi Feng, and Bingyi Kang.
\newblock Video depth anything: Consistent depth estimation for super-long videos.
\newblock In \emph{Proceedings of the Computer Vision and Pattern Recognition Conference}, pp.\  22831--22840, 2025.

\bibitem[Chung et~al.(2024)Chung, Hyun, and Heo]{chung2024style}
Jiwoo Chung, Sangeek Hyun, and Jae-Pil Heo.
\newblock Style injection in diffusion: A training-free approach for adapting large-scale diffusion models for style transfer.
\newblock In \emph{Proceedings of the IEEE/CVF conference on computer vision and pattern recognition}, pp.\  8795--8805, 2024.

\bibitem[Dalal et~al.(2025)Dalal, Koceja, Hussein, Xu, Zhao, Song, Han, Cheung, Kautz, Guestrin, et~al.]{dalal2025one}
Karan Dalal, Daniel Koceja, Gashon Hussein, Jiarui Xu, Yue Zhao, Youjin Song, Shihao Han, Ka~Chun Cheung, Jan Kautz, Carlos Guestrin, et~al.
\newblock One-minute video generation with test-time training.
\newblock \emph{arXiv preprint arXiv:2504.05298}, 2025.

\bibitem[Dao(2024)]{dao2024flashattention}
Tri Dao.
\newblock Flashattention-2: Faster attention with better parallelism and work partitioning.
\newblock In \emph{The Twelfth International Conference on Learning Representations}, 2024.
\newblock URL \url{https://openreview.net/forum?id=mZn2Xyh9Ec}.

\bibitem[Deng et~al.(2022)Deng, Tang, Dong, Ma, Pan, Wang, and Xu]{deng2022stytr2}
Yingying Deng, Fan Tang, Weiming Dong, Chongyang Ma, Xingjia Pan, Lei Wang, and Changsheng Xu.
\newblock Stytr2: Image style transfer with transformers.
\newblock In \emph{Proceedings of the IEEE/CVF conference on computer vision and pattern recognition}, pp.\  11326--11336, 2022.

\bibitem[Deng et~al.(2024)Deng, Wang, Zhang, Tai, and Tang]{deng2024dragvideo}
Yufan Deng, Ruida Wang, Yuhao Zhang, Yu-Wing Tai, and Chi-Keung Tang.
\newblock Dragvideo: Interactive drag-style video editing.
\newblock In \emph{European Conference on Computer Vision}, pp.\  183--199. Springer, 2024.

\bibitem[Duan et~al.(2023)Duan, Wang, Chen, Qian, Huang, and Jin]{duan2023fastblend}
Zhongjie Duan, Chengyu Wang, Cen Chen, Weining Qian, Jun Huang, and Mingyi Jin.
\newblock Fastblend: a powerful model-free toolkit making video stylization easier.
\newblock \emph{arXiv preprint arXiv:2311.09265}, 2023.

\bibitem[Feng et~al.(2024)Feng, Weng, Wang, Yuan, Bao, Luo, Chen, and Guo]{feng2024ccedit}
Ruoyu Feng, Wenming Weng, Yanhui Wang, Yuhui Yuan, Jianmin Bao, Chong Luo, Zhibo Chen, and Baining Guo.
\newblock Ccedit: Creative and controllable video editing via diffusion models.
\newblock In \emph{Proceedings of the IEEE/CVF Conference on Computer Vision and Pattern Recognition}, pp.\  6712--6722, 2024.

\bibitem[Fr{\"u}hst{\"u}ck et~al.(2023)Fr{\"u}hst{\"u}ck, Sarafianos, Xu, Wonka, and Tung]{fruhstuck2023vive3d}
Anna Fr{\"u}hst{\"u}ck, Nikolaos Sarafianos, Yuanlu Xu, Peter Wonka, and Tony Tung.
\newblock Vive3d: Viewpoint-independent video editing using 3d-aware gans.
\newblock In \emph{Proceedings of the IEEE/CVF Conference on Computer Vision and Pattern Recognition}, pp.\  4446--4455, 2023.

\bibitem[Gatys et~al.(2016)Gatys, Ecker, and Bethge]{gatys2016image}
Leon~A Gatys, Alexander~S Ecker, and Matthias Bethge.
\newblock Image style transfer using convolutional neural networks.
\newblock In \emph{Proceedings of the IEEE conference on computer vision and pattern recognition}, pp.\  2414--2423, 2016.

\bibitem[Geyer et~al.(2024)Geyer, Bar-Tal, Bagon, and Dekel]{geyer2024tokenflow}
Michal Geyer, Omer Bar-Tal, Shai Bagon, and Tali Dekel.
\newblock Tokenflow: Consistent diffusion features for consistent video editing.
\newblock In \emph{The Twelfth International Conference on Learning Representations}, 2024.
\newblock URL \url{https://openreview.net/forum?id=lKK50q2MtV}.

\bibitem[Guo et~al.(2024)Guo, Yang, Rao, Liang, Wang, Qiao, Agrawala, Lin, and Dai]{guo2024animatediff}
Yuwei Guo, Ceyuan Yang, Anyi Rao, Zhengyang Liang, Yaohui Wang, Yu~Qiao, Maneesh Agrawala, Dahua Lin, and Bo~Dai.
\newblock Animatediff: Animate your personalized text-to-image diffusion models without specific tuning.
\newblock In \emph{The Twelfth International Conference on Learning Representations}, 2024.
\newblock URL \url{https://openreview.net/forum?id=Fx2SbBgcte}.

\bibitem[Guo et~al.(2025)Guo, Yang, Rao, Meng, Bar-Tal, Ding, Agrawala, Lin, and Dai]{guo2025keyframe}
Yuwei Guo, Ceyuan Yang, Anyi Rao, Chenlin Meng, Omer Bar-Tal, Shuangrui Ding, Maneesh Agrawala, Dahua Lin, and Bo~Dai.
\newblock Keyframe-guided creative video inpainting.
\newblock In \emph{Proceedings of the Computer Vision and Pattern Recognition Conference}, pp.\  13009--13020, 2025.

\bibitem[Hertz et~al.(2022)Hertz, Mokady, Tenenbaum, Aberman, Pritch, and Cohen-Or]{hertz2022prompt}
Amir Hertz, Ron Mokady, Jay Tenenbaum, Kfir Aberman, Yael Pritch, and Daniel Cohen-Or.
\newblock Prompt-to-prompt image editing with cross attention control.
\newblock \emph{arXiv preprint arXiv:2208.01626}, 2022.

\bibitem[Ho et~al.(2020)Ho, Jain, and Abbeel]{ddpm}
Jonathan Ho, Ajay Jain, and Pieter Abbeel.
\newblock Denoising diffusion probabilistic models.
\newblock \emph{Advances in neural information processing systems}, 33:\penalty0 6840--6851, 2020.

\bibitem[Huang \& Belongie(2017)Huang and Belongie]{huang2017arbitrary}
Xun Huang and Serge Belongie.
\newblock Arbitrary style transfer in real-time with adaptive instance normalization.
\newblock In \emph{Proceedings of the IEEE international conference on computer vision}, pp.\  1501--1510, 2017.

\bibitem[Huang et~al.(2024)Huang, He, Yu, Zhang, Si, Jiang, Zhang, Wu, Jin, Chanpaisit, et~al.]{huang2024vbench}
Ziqi Huang, Yinan He, Jiashuo Yu, Fan Zhang, Chenyang Si, Yuming Jiang, Yuanhan Zhang, Tianxing Wu, Qingyang Jin, Nattapol Chanpaisit, et~al.
\newblock Vbench: Comprehensive benchmark suite for video generative models.
\newblock In \emph{Proceedings of the IEEE/CVF Conference on Computer Vision and Pattern Recognition}, pp.\  21807--21818, 2024.

\bibitem[Jiang et~al.(2025)Jiang, Han, Mao, Zhang, Pan, and Liu]{jiang2025vace}
Zeyinzi Jiang, Zhen Han, Chaojie Mao, Jingfeng Zhang, Yulin Pan, and Yu~Liu.
\newblock Vace: All-in-one video creation and editing.
\newblock \emph{arXiv preprint arXiv:2503.07598}, 2025.

\bibitem[Ju et~al.(2024)Ju, Zeng, Bian, Liu, and Xu]{ju2024pnp}
Xuan Ju, Ailing Zeng, Yuxuan Bian, Shaoteng Liu, and Qiang Xu.
\newblock Pnp inversion: Boosting diffusion-based editing with 3 lines of code.
\newblock In \emph{The Twelfth International Conference on Learning Representations}, 2024.
\newblock URL \url{https://openreview.net/forum?id=FoMZ4ljhVw}.

\bibitem[Kahatapitiya et~al.(2024)Kahatapitiya, Liu, He, Liu, Jia, Zhang, Ryoo, and Xie]{kahatapitiya2024adaptive}
Kumara Kahatapitiya, Haozhe Liu, Sen He, Ding Liu, Menglin Jia, Chenyang Zhang, Michael~S Ryoo, and Tian Xie.
\newblock Adaptive caching for faster video generation with diffusion transformers.
\newblock \emph{arXiv preprint arXiv:2411.02397}, 2024.

\bibitem[Kong et~al.(2024)Kong, Tian, Zhang, Min, Dai, Zhou, Xiong, Li, Wu, Zhang, et~al.]{kong2024hunyuanvideo}
Weijie Kong, Qi~Tian, Zijian Zhang, Rox Min, Zuozhuo Dai, Jin Zhou, Jiangfeng Xiong, Xin Li, Bo~Wu, Jianwei Zhang, et~al.
\newblock Hunyuanvideo: A systematic framework for large video generative models.
\newblock \emph{arXiv preprint arXiv:2412.03603}, 2024.

\bibitem[Ku et~al.(2024)Ku, Wei, Ren, Yang, and Chen]{ku2024anyv2v}
Max Ku, Cong Wei, Weiming Ren, Harry Yang, and Wenhu Chen.
\newblock Anyv2v: A tuning-free framework for any video-to-video editing tasks.
\newblock \emph{arXiv preprint arXiv:2403.14468}, 2024.

\bibitem[Li et~al.(2024)Li, Ma, Yang, and Yang]{li2024vidtome}
Xirui Li, Chao Ma, Xiaokang Yang, and Ming-Hsuan Yang.
\newblock Vidtome: Video token merging for zero-shot video editing.
\newblock In \emph{Proceedings of the IEEE/CVF Conference on Computer Vision and Pattern Recognition}, pp.\  7486--7495, 2024.

\bibitem[Lipman et~al.(2023)Lipman, Chen, Ben-Hamu, Nickel, and Le]{flow_matching}
Yaron Lipman, Ricky T.~Q. Chen, Heli Ben-Hamu, Maximilian Nickel, and Matthew Le.
\newblock Flow matching for generative modeling.
\newblock In \emph{The Eleventh International Conference on Learning Representations}, 2023.
\newblock URL \url{https://openreview.net/forum?id=PqvMRDCJT9t}.

\bibitem[Liu et~al.(2024{\natexlab{a}})Liu, Xia, Zhang, Chen, Xing, Wang, Wang, Shan, and Yang]{liu2024stylecrafter}
Gongye Liu, Menghan Xia, Yong Zhang, Haoxin Chen, Jinbo Xing, Yibo Wang, Xintao Wang, Ying Shan, and Yujiu Yang.
\newblock Stylecrafter: Taming artistic video diffusion with reference-augmented adapter learning.
\newblock \emph{ACM Trans. Graph.}, 2024{\natexlab{a}}.

\bibitem[Liu et~al.(2024{\natexlab{b}})Liu, Zhang, Li, Lin, and Jia]{liu2024video}
Shaoteng Liu, Yuechen Zhang, Wenbo Li, Zhe Lin, and Jiaya Jia.
\newblock Video-p2p: Video editing with cross-attention control.
\newblock In \emph{Proceedings of the IEEE/CVF Conference on Computer Vision and Pattern Recognition}, pp.\  8599--8608, 2024{\natexlab{b}}.

\bibitem[Liu et~al.(2021)Liu, Lin, He, Li, Wang, Li, Sun, Li, and Ding]{liu2021adaattn}
Songhua Liu, Tianwei Lin, Dongliang He, Fu~Li, Meiling Wang, Xin Li, Zhengxing Sun, Qian Li, and Errui Ding.
\newblock Adaattn: Revisit attention mechanism in arbitrary neural style transfer.
\newblock In \emph{Proceedings of the IEEE/CVF international conference on computer vision}, pp.\  6649--6658, 2021.

\bibitem[Ma et~al.(2025)Ma, Cun, Liang, Xing, He, Qi, Chen, and Chen]{ma2025magicstick}
Yue Ma, Xiaodong Cun, Sen Liang, Jinbo Xing, Yingqing He, Chenyang Qi, Siran Chen, and Qifeng Chen.
\newblock Magicstick: Controllable video editing via control handle transformations.
\newblock In \emph{2025 IEEE/CVF Winter Conference on Applications of Computer Vision (WACV)}, pp.\  9385--9395. IEEE, 2025.

\bibitem[Mao et~al.(2024)Mao, Huang, Xie, Chang, Hui, Xu, and Zhou]{mao2024story}
Jiawei Mao, Xiaoke Huang, Yunfei Xie, Yuanqi Chang, Mude Hui, Bingjie Xu, and Yuyin Zhou.
\newblock Story-adapter: A training-free iterative framework for long story visualization.
\newblock \emph{arXiv preprint arXiv:2410.06244}, 2024.

\bibitem[Mei \& Patel(2023)Mei and Patel]{mei2023}
Kangfu Mei and Vishal Patel.
\newblock Vidm: video implicit diffusion models.
\newblock In \emph{Proceedings of the Thirty-Seventh AAAI Conference on Artificial Intelligence and Thirty-Fifth Conference on Innovative Applications of Artificial Intelligence and Thirteenth Symposium on Educational Advances in Artificial Intelligence}, AAAI'23/IAAI'23/EAAI'23. AAAI Press, 2023.
\newblock ISBN 978-1-57735-880-0.
\newblock \doi{10.1609/aaai.v37i8.26094}.
\newblock URL \url{https://doi.org/10.1609/aaai.v37i8.26094}.

\bibitem[Ouyang et~al.(2024)Ouyang, Dong, Yang, Si, and Pan]{ouyang2024i2vedit}
Wenqi Ouyang, Yi~Dong, Lei Yang, Jianlou Si, and Xingang Pan.
\newblock I2vedit: First-frame-guided video editing via image-to-video diffusion models.
\newblock In \emph{SIGGRAPH Asia 2024 Conference Papers}, pp.\  1--11, 2024.

\bibitem[Pande \& Sharma(2023)Pande and Sharma]{pande2023vist3d}
Ayush Pande and Gaurav Sharma.
\newblock Vist3d: video stylization with 3d cnn.
\newblock \emph{Advances in Neural Information Processing Systems}, 36:\penalty0 41651--41662, 2023.

\bibitem[Parmar et~al.(2023)Parmar, Kumar~Singh, Zhang, Li, Lu, and Zhu]{parmar2023zero}
Gaurav Parmar, Krishna Kumar~Singh, Richard Zhang, Yijun Li, Jingwan Lu, and Jun-Yan Zhu.
\newblock Zero-shot image-to-image translation.
\newblock In \emph{ACM SIGGRAPH 2023 conference proceedings}, pp.\  1--11, 2023.

\bibitem[Peebles \& Xie(2023)Peebles and Xie]{peebles2023scalable}
William Peebles and Saining Xie.
\newblock Scalable diffusion models with transformers.
\newblock In \emph{Proceedings of the IEEE/CVF international conference on computer vision}, pp.\  4195--4205, 2023.

\bibitem[Radford et~al.(2021)Radford, Kim, Hallacy, Ramesh, Goh, Agarwal, Sastry, Askell, Mishkin, Clark, et~al.]{radford2021learning}
Alec Radford, Jong~Wook Kim, Chris Hallacy, Aditya Ramesh, Gabriel Goh, Sandhini Agarwal, Girish Sastry, Amanda Askell, Pamela Mishkin, Jack Clark, et~al.
\newblock Learning transferable visual models from natural language supervision.
\newblock In \emph{International conference on machine learning}, pp.\  8748--8763. PmLR, 2021.

\bibitem[Ren et~al.(2025)Ren, Shen, Huang, Ling, Lu, Nimier-David, M{\"u}ller, Keller, Fidler, and Gao]{ren2025gen3c}
Xuanchi Ren, Tianchang Shen, Jiahui Huang, Huan Ling, Yifan Lu, Merlin Nimier-David, Thomas M{\"u}ller, Alexander Keller, Sanja Fidler, and Jun Gao.
\newblock Gen3c: 3d-informed world-consistent video generation with precise camera control.
\newblock \emph{arXiv preprint arXiv:2503.03751}, 2025.

\bibitem[Rout et~al.(2025)Rout, Chen, Ruiz, Kumar, Caramanis, Shakkottai, and Chu]{rout2025rb}
Litu Rout, Yujia Chen, Nataniel Ruiz, Abhishek Kumar, Constantine Caramanis, Sanjay Shakkottai, and Wen-Sheng Chu.
\newblock Rb-modulation: Training-free stylization using reference-based modulation.
\newblock In \emph{The Thirteenth International Conference on Learning Representations}, 2025.

\bibitem[Sim{\'e}oni et~al.(2025)Sim{\'e}oni, Vo, Seitzer, Baldassarre, Oquab, Jose, Khalidov, Szafraniec, Yi, Ramamonjisoa, et~al.]{simeoni2025dinov3}
Oriane Sim{\'e}oni, Huy~V Vo, Maximilian Seitzer, Federico Baldassarre, Maxime Oquab, Cijo Jose, Vasil Khalidov, Marc Szafraniec, Seungeun Yi, Micha{\"e}l Ramamonjisoa, et~al.
\newblock Dinov3.
\newblock \emph{arXiv preprint arXiv:2508.10104}, 2025.

\bibitem[Somepalli et~al.(2024)Somepalli, Gupta, Gupta, Palta, Goldblum, Geiping, Shrivastava, and Goldstein]{somepalli2024measuring}
Gowthami Somepalli, Anubhav Gupta, Kamal Gupta, Shramay Palta, Micah Goldblum, Jonas Geiping, Abhinav Shrivastava, and Tom Goldstein.
\newblock Measuring style similarity in diffusion models.
\newblock \emph{arXiv preprint arXiv:2404.01292}, 2024.

\bibitem[Staniszewski et~al.(2024)Staniszewski, Kuci{\'n}ski, and Deja]{staniszewski2024there}
{\L}ukasz Staniszewski, {\L}ukasz Kuci{\'n}ski, and Kamil Deja.
\newblock There and back again: On the relation between noise and image inversions in diffusion models.
\newblock \emph{arXiv preprint arXiv:2410.23530}, 2024.

\bibitem[Tan et~al.(2018)Tan, Chan, Aguirre, and Tanaka]{tan2018improved}
Wei~Ren Tan, Chee~Seng Chan, Hernan~E Aguirre, and Kiyoshi Tanaka.
\newblock Improved artgan for conditional synthesis of natural image and artwork.
\newblock \emph{IEEE Transactions on Image Processing}, 28\penalty0 (1):\penalty0 394--409, 2018.

\bibitem[Teed \& Deng(2020)Teed and Deng]{teed2020raft}
Zachary Teed and Jia Deng.
\newblock Raft: Recurrent all-pairs field transforms for optical flow.
\newblock In \emph{European conference on computer vision}, pp.\  402--419. Springer, 2020.

\bibitem[Tu et~al.(2025)Tu, Luo, Chen, Ji, Bai, and Zhao]{tu2025videoanydoor}
Yuanpeng Tu, Hao Luo, Xi~Chen, Sihui Ji, Xiang Bai, and Hengshuang Zhao.
\newblock Videoanydoor: High-fidelity video object insertion with precise motion control.
\newblock In \emph{Proceedings of the Special Interest Group on Computer Graphics and Interactive Techniques Conference Conference Papers}, pp.\  1--11, 2025.

\bibitem[Tzaban et~al.(2022)Tzaban, Mokady, Gal, Bermano, and Cohen-Or]{tzaban2022stitch}
Rotem Tzaban, Ron Mokady, Rinon Gal, Amit Bermano, and Daniel Cohen-Or.
\newblock Stitch it in time: Gan-based facial editing of real videos.
\newblock In \emph{SIGGRAPH Asia 2022 Conference Papers}, pp.\  1--9, 2022.

\bibitem[Wan et~al.(2025)Wan, Wang, Ai, Wen, Mao, Xie, Chen, Yu, Zhao, Yang, et~al.]{wan2025wan}
Team Wan, Ang Wang, Baole Ai, Bin Wen, Chaojie Mao, Chen-Wei Xie, Di~Chen, Feiwu Yu, Haiming Zhao, Jianxiao Yang, et~al.
\newblock Wan: Open and advanced large-scale video generative models.
\newblock \emph{arXiv preprint arXiv:2503.20314}, 2025.

\bibitem[Wang et~al.(2024{\natexlab{a}})Wang, Spinelli, Wang, Bai, Qin, and Chen]{wang2024instantstyle}
Haofan Wang, Matteo Spinelli, Qixun Wang, Xu~Bai, Zekui Qin, and Anthony Chen.
\newblock Instantstyle: Free lunch towards style-preserving in text-to-image generation.
\newblock \emph{arXiv preprint arXiv:2404.02733}, 2024{\natexlab{a}}.

\bibitem[Wang et~al.(2024{\natexlab{b}})Wang, Xing, Huang, Ai, Wang, and Bai]{wang2024instantstyle_plus}
Haofan Wang, Peng Xing, Renyuan Huang, Hao Ai, Qixun Wang, and Xu~Bai.
\newblock Instantstyle-plus: Style transfer with content-preserving in text-to-image generation.
\newblock \emph{arXiv preprint arXiv:2407.00788}, 2024{\natexlab{b}}.

\bibitem[Wang et~al.(2020)Wang, Yang, Xu, and Liu]{wang2020consistent}
Wenjing Wang, Shuai Yang, Jizheng Xu, and Jiaying Liu.
\newblock Consistent video style transfer via relaxation and regularization.
\newblock \emph{IEEE Transactions on Image Processing}, 29:\penalty0 9125--9139, 2020.

\bibitem[Wang et~al.(2023)Wang, Zhao, and Xing]{wang2023stylediffusion}
Zhizhong Wang, Lei Zhao, and Wei Xing.
\newblock Stylediffusion: Controllable disentangled style transfer via diffusion models.
\newblock In \emph{Proceedings of the IEEE/CVF international conference on computer vision}, pp.\  7677--7689, 2023.

\bibitem[Wright \& Ommer(2022)Wright and Ommer]{wright2022artfid}
Matthias Wright and Bj{\"o}rn Ommer.
\newblock Artfid: Quantitative evaluation of neural style transfer.
\newblock In \emph{DAGM German Conference on Pattern Recognition}, pp.\  560--576. Springer, 2022.

\bibitem[Xi et~al.(2025)Xi, Yang, Zhao, Xu, Li, Li, Lin, Cai, Zhang, Li, et~al.]{xi2025sparse}
Haocheng Xi, Shuo Yang, Yilong Zhao, Chenfeng Xu, Muyang Li, Xiuyu Li, Yujun Lin, Han Cai, Jintao Zhang, Dacheng Li, et~al.
\newblock Sparse videogen: Accelerating video diffusion transformers with spatial-temporal sparsity.
\newblock \emph{arXiv preprint arXiv:2502.01776}, 2025.

\bibitem[Xing et~al.(2024)Xing, Wang, Sun, Wang, Bai, Ai, Huang, and Li]{xing2024csgo}
Peng Xing, Haofan Wang, Yanpeng Sun, Qixun Wang, Xu~Bai, Hao Ai, Renyuan Huang, and Zechao Li.
\newblock Csgo: Content-style composition in text-to-image generation.
\newblock \emph{arXiv preprint arXiv:2408.16766}, 2024.

\bibitem[Xu et~al.(2025{\natexlab{a}})Xu, Jiang, Hu, Luo, He, Zhang, Wang, Wu, Ling, and Wang]{euler_inversion}
Pengcheng Xu, Boyuan Jiang, Xiaobin Hu, Donghao Luo, Qingdong He, Jiangning Zhang, Chengjie Wang, Yunsheng Wu, Charles Ling, and Boyu Wang.
\newblock Unveil inversion and invariance in flow transformer for versatile image editing.
\newblock In \emph{Proceedings of the Computer Vision and Pattern Recognition Conference}, pp.\  28479--28489, 2025{\natexlab{a}}.

\bibitem[Xu et~al.(2019)Xu, Li, Zhou, and Loy]{xu2019deep}
Rui Xu, Xiaoxiao Li, Bolei Zhou, and Chen~Change Loy.
\newblock Deep flow-guided video inpainting.
\newblock In \emph{Proceedings of the IEEE/CVF conference on computer vision and pattern recognition}, pp.\  3723--3732, 2019.

\bibitem[Xu et~al.(2025{\natexlab{b}})Xu, Xi, Wang, Mao, and Cheng]{xu2025stylessp}
Ruojun Xu, Weijie Xi, XiaoDi Wang, Yongbo Mao, and Zach Cheng.
\newblock Stylessp: Sampling startpoint enhancement for training-free diffusion-based method for style transfer.
\newblock In \emph{Proceedings of the Computer Vision and Pattern Recognition Conference}, pp.\  18260--18269, 2025{\natexlab{b}}.

\bibitem[Yang et~al.(2025{\natexlab{a}})Yang, Li, Yang, Zhang, Hui, Zheng, Yu, Gao, Huang, Lv, et~al.]{yang2025qwen3}
An~Yang, Anfeng Li, Baosong Yang, Beichen Zhang, Binyuan Hui, Bo~Zheng, Bowen Yu, Chang Gao, Chengen Huang, Chenxu Lv, et~al.
\newblock Qwen3 technical report.
\newblock \emph{arXiv preprint arXiv:2505.09388}, 2025{\natexlab{a}}.

\bibitem[Yang et~al.(2025{\natexlab{b}})Yang, Zhu, Fan, and Yang]{yang2025videograin}
Xiangpeng Yang, Linchao Zhu, Hehe Fan, and Yi~Yang.
\newblock Videograin: Modulating space-time attention for multi-grained video editing.
\newblock In \emph{The Thirteenth International Conference on Learning Representations}, 2025{\natexlab{b}}.
\newblock URL \url{https://openreview.net/forum?id=SSslAtcPB6}.

\bibitem[Yang et~al.(2023)Yang, Zhou, Feng, and Wang]{yang2023diffusion}
Xingyi Yang, Daquan Zhou, Jiashi Feng, and Xinchao Wang.
\newblock Diffusion probabilistic model made slim.
\newblock In \emph{Proceedings of the IEEE/CVF Conference on computer vision and pattern recognition}, pp.\  22552--22562, 2023.

\bibitem[Yang et~al.(2025{\natexlab{c}})Yang, Teng, Zheng, Ding, Huang, Xu, Yang, Hong, Zhang, Feng, Yin, Yuxuan.Zhang, Wang, Cheng, Xu, Gu, Dong, and Tang]{yang2025cogvideox}
Zhuoyi Yang, Jiayan Teng, Wendi Zheng, Ming Ding, Shiyu Huang, Jiazheng Xu, Yuanming Yang, Wenyi Hong, Xiaohan Zhang, Guanyu Feng, Da~Yin, Yuxuan.Zhang, Weihan Wang, Yean Cheng, Bin Xu, Xiaotao Gu, Yuxiao Dong, and Jie Tang.
\newblock Cogvideox: Text-to-video diffusion models with an expert transformer.
\newblock In \emph{The Thirteenth International Conference on Learning Representations}, 2025{\natexlab{c}}.
\newblock URL \url{https://openreview.net/forum?id=LQzN6TRFg9}.

\bibitem[Ye et~al.(2025)Ye, Huang, Wang, Wan, Zhang, and Luo]{ye2025stylemaster}
Zixuan Ye, Huijuan Huang, Xintao Wang, Pengfei Wan, Di~Zhang, and Wenhan Luo.
\newblock Stylemaster: Stylize your video with artistic generation and translation.
\newblock In \emph{Proceedings of the Computer Vision and Pattern Recognition Conference}, pp.\  2630--2640, 2025.

\bibitem[Zhang et~al.(2025{\natexlab{a}})Zhang, Li, Chen, Ge, Sun, Zhang, Jiang, Yuan, Peng, and Luo]{zhang2025flashvideo}
Shilong Zhang, Wenbo Li, Shoufa Chen, Chongjian Ge, Peize Sun, Yida Zhang, Yi~Jiang, Zehuan Yuan, Binyue Peng, and Ping Luo.
\newblock Flashvideo: Flowing fidelity to detail for efficient high-resolution video generation.
\newblock \emph{arXiv preprint arXiv:2502.05179}, 2025{\natexlab{a}}.

\bibitem[Zhang et~al.(2025{\natexlab{b}})Zhang, Zhou, Qin, Lu, Yan, Wang, Chen, and Liu]{zhang2025enabling}
Xu~Zhang, Hao Zhou, Haoming Qin, Xiaobin Lu, Jiaxing Yan, Guanzhong Wang, Zeyu Chen, and Yi~Liu.
\newblock Enabling versatile controls for video diffusion models.
\newblock \emph{arXiv preprint arXiv:2503.16983}, 2025{\natexlab{b}}.

\bibitem[Zhang et~al.(2025{\natexlab{c}})Zhang, Ju, and Clark]{zhang2025fastvideoedit}
Youyuan Zhang, Xuan Ju, and James~J Clark.
\newblock Fastvideoedit: Leveraging consistency models for efficient text-to-video editing.
\newblock In \emph{2025 IEEE/CVF Winter Conference on Applications of Computer Vision (WACV)}, pp.\  3657--3666. IEEE, 2025{\natexlab{c}}.

\bibitem[Zhang et~al.(2023)Zhang, Huang, Tang, Huang, Ma, Dong, and Xu]{zhang2023inversion}
Yuxin Zhang, Nisha Huang, Fan Tang, Haibin Huang, Chongyang Ma, Weiming Dong, and Changsheng Xu.
\newblock Inversion-based style transfer with diffusion models.
\newblock In \emph{Proceedings of the IEEE/CVF conference on computer vision and pattern recognition}, pp.\  10146--10156, 2023.

\bibitem[Zhao et~al.(2025)Zhao, He, Chen, Zhu, Li, and Zhu]{zhao2025riflex}
Min Zhao, Guande He, Yixiao Chen, Hongzhou Zhu, Chongxuan Li, and Jun Zhu.
\newblock Riflex: A free lunch for length extrapolation in video diffusion transformers.
\newblock \emph{arXiv preprint arXiv:2502.15894}, 2025.

\end{thebibliography}
\bibliographystyle{iclr2026_conference}

\newpage
\appendix
\section{Appendix}
\subsection{Preliminary: PnP Inversion}
\begin{figure}[!t]  
\vspace{-2ex}  
\includegraphics[width=\linewidth]{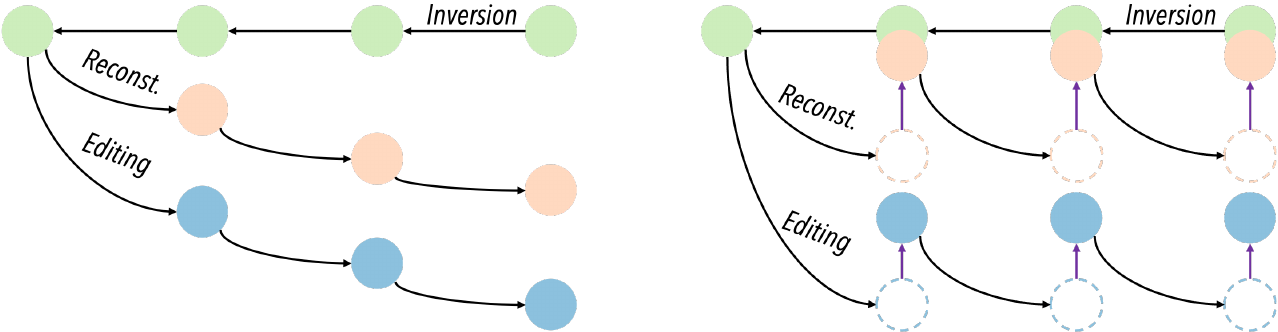}
\caption{
Latent movement along the inversion and denoising process for reconstruction and editing. Left: Conventional Inversion; Right: PnP Inversion. Purple arrows refer to the compensation.
}
\label{fig:pnp}
\vspace{-2ex}  
\end{figure}
Different from conventional inversion-based editing methods, PnP inversion~\cite{ju2024pnp} records the full denoising trajectory 
\([\,\mathbf{x}_{t}, \mathbf{x}_{t-1}, \ldots, \mathbf{x}_{0}\,]\) (the \emph{target latents}) during inversion, and explicitly injects the discrepancy between the reconstruction latent \(\mathbf{x}^{r}_{t}\) and the target latent \(\mathbf{x}_{t}\) into both the reconstruction and editing latents \(\mathbf{x}^{e}_{t}\). 
This compensation mitigates the accumulated error that is otherwise amplified by classifier-free guidance, thereby steering both reconstruction and editing back toward the original denoising trajectory (Figure~\ref{fig:pnp}). 
The update rule can be expressed as:
\begin{equation}
    \mathbf{x}^{r}_{t} \leftarrow \mathbf{x}^{r}_{t} + \big(\mathbf{x}_{t} - \mathbf{x}^{r}_{t}\big), 
    \quad 
    \mathbf{x}^{e}_{t} \leftarrow \mathbf{x}^{e}_{t} + \big(\mathbf{x}_{t} - \mathbf{x}^{r}_{t}\big)
\end{equation}
In this paper, we observe that the inverted noise alone is insufficient to guarantee the correct reconstruction of the layout and motion in modern video diffusion models, and PnP inversion (combined with our IHC strategy) can successfully resolve this issue.

\subsection{Implementation Details}
Control signals such as depth, Canny edges, and tile images offer additional content guidance for generative models and have been widely adopted in both training-free and training-based style transfer approaches \cite{xu2025stylessp, xing2024csgo, wang2024instantstyle, ye2025stylemaster}. To enhance motion preservation, we adopt Wan2.1-Fun-Control as our pretrained base model, an adapted variant of Wan2.1 \cite{wan2025wan} that can incorporate depth as a control signal. This base model positions the reference tokens at the beginning of the visual token sequence and supports only a single reference image, which serves as the initial frame of the generated video. One can freely change the base model to other video diffusion models, given the FreeViS framework. We utilize Euler Inversion with fixed-point iteration \cite{euler_inversion} to perform inversion under the Rectified Flow framework \cite{flow_matching}. The number of iterations can be adjusted based on the users' computational budgets. During inversion, we omit the reference latent and provide only the video depth map parsed by Video Depth Anything \cite{chen2025video} and a concise text prompt extracted using Qwen2.5-VL \cite{bai2025qwen25vltechnicalreport}, which describes the content and dynamics of the input video. 

\subsection{Computation of Reference Masks}
\label{app:ref_mask}

Let video frames be indexed by $t=0,\dots,T-1$ with pixel domain $\Omega_t\subset\mathbb{Z}^2$. We adopt RAFT \cite{teed2020raft} $\mathbf{f}_t:\mathbb{R}^2\to\mathbb{R}^2$ to extract the forward optical flow from frame $t$ to $t{+}1$, and let $\Pi:\mathbb{R}^2\to\Omega_t$ denote nearest-neighbor discretization to pixels.
\begin{equation}
\label{eq:propagation}
\mathbf{p}_s=\mathbf{p},\quad 
\mathbf{p}_{k+1}=\mathbf{p}_{k}+\mathbf{f}_{k}\!\big(\mathbf{p}_{k}\big)\quad (k=s,\dots,t{-}1), 
\qquad
\Phi_{s\to t}(\mathbf{p})=\mathbf{p}_t
\end{equation}
where $\Phi_{s\to t}$ denotes the forward propagation operator. Given a target frame $t$ and a source set $S\subseteq\{0,\dots,T{-}1\}$, the set of pixels in frame $t$ covered by $S$ is computed by:
\begin{equation}
\label{eq:coverage}
\mathcal{C}_t(S)=\Big\{\;\Pi\!\big(\Phi_{a\to t}(\mathbf{u})\big)\;:\;\mathbf{u}\in\Omega_a,\; a\in S\;\Big\}\subseteq\Omega_t
\end{equation}
With two given frame indices $i$ and $j$ ($0< i<j\le T-1$), the two novel-region masks are:
\begin{equation}
\label{eq:masks}
M_{i}(\mathbf{p})=\mathds{1}\!\big[\mathbf{p}\notin \mathcal{C}_{i}(\{0\})\big],
\qquad
M_{j}(\mathbf{p})=\mathds{1}\!\big[\mathbf{p}\notin \mathcal{C}_{j}(\{0,i\})\big],
\quad \mathbf{p}\in\Omega_t
\end{equation}
This approach can be extended to more references with a pre-defined order.

\subsection{Computation of Flow Masks}
\label{app:flow_mask}
We construct a dense binary correspondence mask that records, for every source pixel in frame $s$, the (discretized) terminal location it reaches in any target frame $t$ when transported by optical flow. Let frames be indexed by $t=0,\dots,T{-}1$ with pixel domains $\Omega_t\subset\mathbb{Z}^2$. As in \S\ref{app:ref_mask}, RAFT \cite{teed2020raft} provides forward flows $\mathbf{f}_k:\mathbb{R}^2\!\to\!\mathbb{R}^2$ from $k$ to $k{+}1$ (and analogously backward flows $\tilde{\mathbf{f}}_k$ from $k{+}1$ to $k$), and $\Pi_t:\mathbb{R}^2\!\to\!\Omega_t$ denotes nearest-neighbor discretization onto frame $t$.

\paragraph{Intermediate sampling and terminal validity.}
Starting at $\mathbf{p}_s=\mathbf{u}\in\mathbb{R}^2$, we propagate \emph{forward} from $s$ to $t\ge s$ by sampling the flow at the nearest valid pixel and updating the \emph{continuous} state:
\begin{equation}
\label{eq:forward-comp}
\hat{\mathbf{p}}_k=\Pi_k(\mathbf{p}_k),\qquad
\mathbf{p}_{k+1}=\mathbf{p}_k+\mathbf{f}_k\!\big(\hat{\mathbf{p}}_k\big),
\quad k=s,\dots,t{-}1,
\qquad
\Phi_{s\to t}(\mathbf{u})=\mathbf{p}_t
\end{equation}
The \emph{backward} transport for $t<s$ is defined analogously using $\tilde{\mathbf{f}}_k$:
\begin{equation}
\label{eq:backward-comp}
\hat{\mathbf{p}}_{k+1}=\Pi_{k+1}(\mathbf{p}_{k+1}),\qquad
\mathbf{p}_{k}=\mathbf{p}_{k+1}+\tilde{\mathbf{f}}_{k}\!\big(\hat{\mathbf{p}}_{k+1}\big),
\quad k=s{-}1,\dots,t,
\qquad
\tilde{\Phi}_{s\to t}(\mathbf{u})=\mathbf{p}_t
\end{equation}
Note that intermediate positions $\mathbf{p}_k$ are allowed to leave the image domain; $\Pi_k(\cdot)$ is used \emph{only} to fetch subsequent flow values. We enforce validity \emph{solely at the terminal point}: a trajectory from $s$ to $t$ is considered valid iff the continuous endpoint lies in $\Omega_t$.

\paragraph{Bidirectional map and landing index.}
Combining \eqref{eq:forward-comp}--\eqref{eq:backward-comp}, define the bidirectional transport
\begin{equation}
\label{eq:lambda}
\Lambda_{s\to t}(\mathbf{u}) \,=\,
\begin{cases}
\Phi_{s\to t}(\mathbf{u}), & t\ge s,\\[2pt]
\tilde{\Phi}_{s\to t}(\mathbf{u}), & t< s
\end{cases}
\end{equation}
Given $\mathbf{u}\in\Omega_s$, its terminal continuous location is $\mathbf{q}=\Lambda_{s\to t}(\mathbf{u})$, and its discretized landing pixel is
\begin{equation}
\label{eq:landing}
\mathbf{v}^\star=\Pi_t\!\big(\mathbf{q}\big)\in\Omega_t,
\qquad
\chi_{s\to t}(\mathbf{u})=\mathds{1}\!\big[\mathbf{q}\in\Omega_t\big]\in\{0,1\}
\end{equation}

\paragraph{Pairwise mask and tensor stacking.}
The pairwise correspondence mask $M_{s\to t}:\Omega_s\times\Omega_t\to\{0,1\}$ places a single one at the discretized landing pixel when the terminal point is in-bounds:
\begin{equation}
\label{eq:pair-mask}
M_{s\to t}(\mathbf{u},\mathbf{v})
\;=\;
\chi_{s\to t}(\mathbf{u})\;\mathds{1}\!\big[\mathbf{v}=\mathbf{v}^\star\big],
\qquad \mathbf{u}\in\Omega_s,\ \mathbf{v}\in\Omega_t
\end{equation}
Consequently, $\sum_{\mathbf{v}\in\Omega_t} M_{s\to t}(\mathbf{u},\mathbf{v})\in\{0,1\}$ for every $\mathbf{u}$, and the self-map reduces to the pixelwise identity: $M_{s\to s}(\mathbf{u},\mathbf{v})=\mathds{1}[\mathbf{v}=\mathbf{u}]$. Stacking $M_{s\to t}$ over all $(s,t)$ yields the rank-6 tensor
\begin{equation}
\label{eq:rank6}
M\in\{0,1\}^{T\times h\times w\times T\times h\times w},
\qquad
M[s,y,x,t,y',x']=M_{s\to t}\big((y,x),(y',x')\big)
\end{equation}

\paragraph{Remarks.}
In practice we adopt nearest-neighbor sampling via $\Pi_t(\cdot)$ in \eqref{eq:forward-comp}–\eqref{eq:backward-comp}; bilinear interpolation can be substituted without changing the definition in \eqref{eq:pair-mask} because the landing index is discretized by $\Pi_t(\cdot)$. Our construction matches the implementation choice that \emph{intermediate} out-of-bounds positions are permitted, while \emph{only} the \emph{terminal} in-bounds condition $\chi_{s\to t}$ is enforced. Alternatively, the optical flow module can be replaced with other correspondence estimation methods, such as DINOv3 \cite{simeoni2025dinov3} features, for improved performance.

\subsection{Dynamic–Appearance Decomposition for a Causal VAE}
\label{app:decomposition}
\begin{figure}[!t]  
\vspace{-2ex}  
\includegraphics[width=\linewidth]{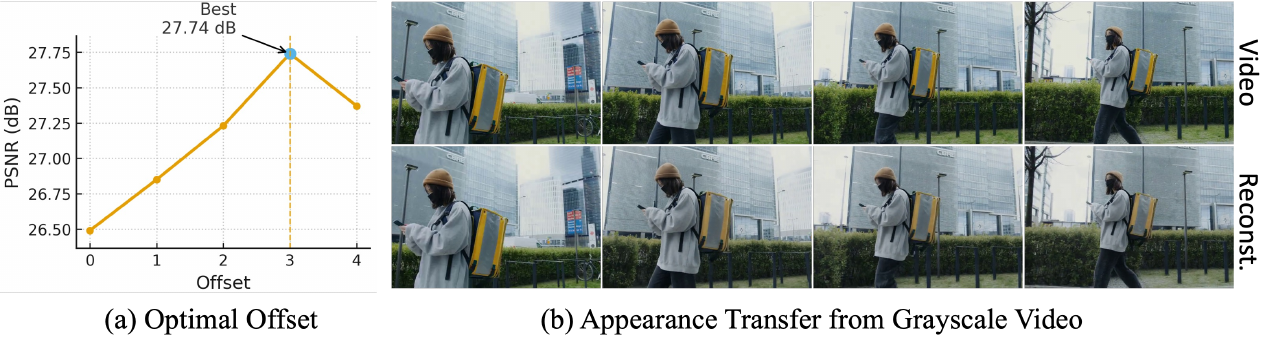}
\caption{
Optimal Offset and visualization for appearance transfer. (a) The plot for reconstruction PSNR with different offsets; (b) The RGB video and its reconstruction via appearance transfer.
}
\label{fig:offset}
\vspace{-2ex}  
\end{figure}

The causal VAE in Wan~\cite{wan2025wan} encodes a video using a temporal subsampling factor \(r\), producing a \emph{time-compressed} latent sequence. We show that this video latent admits a decomposition into (i) an \emph{appearance} term mainly determined by one frame per \(r\)-frame block and (ii) a \emph{dynamic} residual. For \(r=4\), the clip’s appearance is therefore dominated by \(1/4\) of the frames.

\textbf{Sampling and per-frame encoding.}
Given an \(N\)-frame video \(V=\{V[t]\}_{t=0}^{N-1}\) and an offset \(\Delta p\in\{0,\dots,r\}\), define the sampled frame \emph{sequence}
\[
S_r(V,\Delta p)=\big(V[0]\big)\;\cup\;\big(V[rj+\Delta p] \,\big|\, j=0,1,\dots,\lfloor (N-1)/r \rfloor-1\big)
\]
Let \(\mathscr{E}_{\mathrm{vid}}\) be the video encoder and \(\mathscr{E}_{\mathrm{img}}\) the same backbone in image (framewise) mode. The video encoder yields:
\[
z(V)=\mathscr{E}_{\mathrm{vid}}(V)
\]
The latents for sampled sequence is obtained by encoding the sampled frames \emph{independently} and stacking them along time:
\[
a(V,\Delta p)=\big(\,\mathscr{E}_{\mathrm{img}}(I)\,\big)_{I\in S_r(V,\Delta p)}
\]
Because the sampled frames are independently encoded, $a(V,\Delta p)$ only contains appearance information. Define the dynamic residual by:
\[
z_{\mathrm{dyn}}(V,\Delta p)=z(V)-a(V,\Delta p)
\]

\paragraph{Appearance–dynamics exchange.}
For an RGB clip \(V_{\mathrm{rgb}}\) and its grayscale counterpart \(V_{\mathrm{gray}}\), swapping appearance while preserving dynamics gives:
\begin{align}
V_{\mathrm{rgb}}'(\Delta p)&=\mathscr{D}\!\left(z_{\mathrm{dyn}}(V_{\mathrm{gray}},\Delta p)+a(V_{\mathrm{rgb}},\Delta p)\right),\\
V_{\mathrm{gray}}'(\Delta p)&=\mathscr{D}\!\left(z_{\mathrm{dyn}}(V_{\mathrm{rgb}},\Delta p)+a(V_{\mathrm{gray}},\Delta p)\right),
\end{align}
where \(\mathscr{D}\) denotes the VAE decoder. This implements “subtract source appearance, add target appearance” while keeping the dynamics fixed. Figure~\ref{fig:offset}-(b) demonstrates that the appearance of the RGB video is successfully combined with grayscale dynamics, with only a minor color shift. This observation inspires us to inject the dynamic residual into the value matrices of static reference latents, which addresses the stuttering issue of additional reference inputs in Section \ref{sec:references}.

\paragraph{Choosing the offset.}
\(\Delta p\) selects which frame within each \(r\)-frame block anchors appearance. We choose it by minimizing cross-reconstruction error:
\[
\Delta p^\star
=\arg\min_{\Delta p\in\{0,\dots,r\}}
\mathcal{L}\!\left(V_{\mathrm{rgb}}'(\Delta p),V_{\mathrm{rgb}}\right)
+\mathcal{L}\!\left(V_{\mathrm{gray}}'(\Delta p),V_{\mathrm{gray}}\right),
\]
where \(\mathcal{L}\) can be an \(\ell_1\)+LPIPS video reconstruction metric. As shown in Figure~\ref{fig:offset}-(a), the reconstruction performance is optimal when $\Delta p=3$ for $r=4$. Furthermore, the PCA visualizations of the video latents and the sampled frames exhibit close alignment (Figure \ref{fig:rabbit_denoise}), indicating that the temporal anchoring is consistent across the two representations. The additional references should be selected from $S_r(V,\Delta p)$, and the optical flow for inconsistency masking and EOG in Section \ref{sec:references} should be calculated accordingly for optimal performance.

\subsection{Denoising Process of Wan I2V Model}
\label{app:denoising}
\begin{figure}[!t]
    \vspace{-1ex}
    \includegraphics[width=\linewidth]{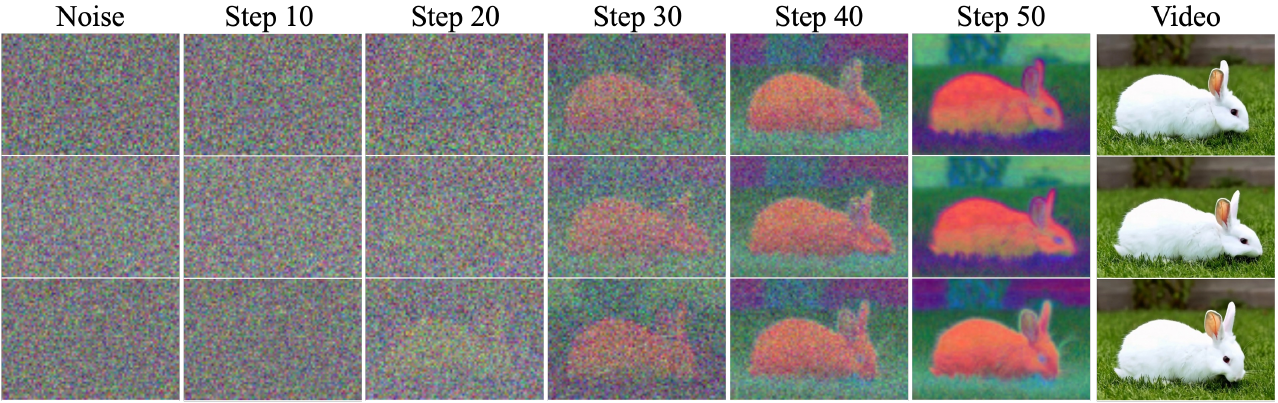}
    \vspace{-4ex}
    \caption{
    Denoising process of Wan I2V model. We apply PCA to visualize the latents for different frames along the entire denoising process. The corresponding sampled video frames are given.
    \vspace{-1ex}
}
    \label{fig:rabbit_denoise}
\end{figure}
As in image diffusion models, the early denoising stages in video diffusion primarily establish the coarse global layout and motion of the generated video, while the later stages are dedicated to refining fine-grained details. As illustrated in Figure \ref{fig:rabbit_denoise}, the primary layout and structure of the entire video are largely determined by around the 20th denoising step. In later steps, the model mainly focuses on the local appearance refinement. The PCA visualizations of the latents align well with our sampled frames with optimal offset.

\begin{figure}  
\vspace{-1ex}  
\includegraphics[width=\linewidth]{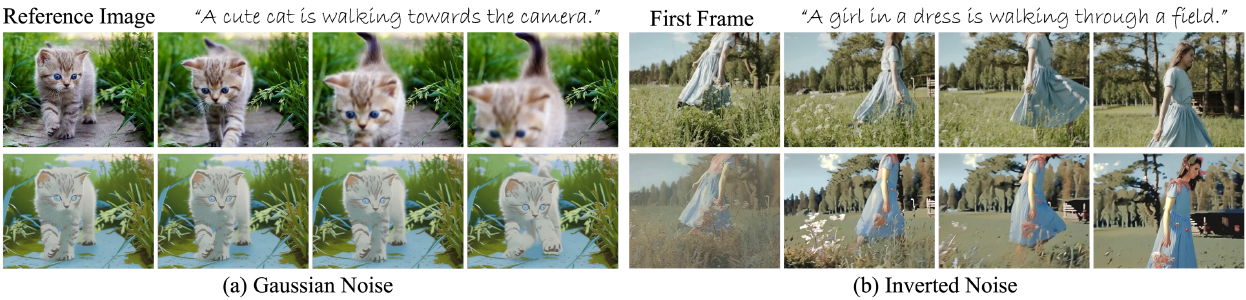}
\caption{
Effect of noise initialization for video stylization with reference. (a) The generated videos share the same initial Gaussian noise. (b) Original video and its stylization via inversion.
}
\label{fig:noise_init}
\vspace{-2ex}  
\end{figure}
\subsection{Noise Initialization}
\label{app:noise}
In recent diffusion-based image and video editing works \cite{hertz2022prompt, cao2023masactrl, parmar2023zero, yang2025videograin, li2024vidtome}, noise inversion is a prerequisite step for essential content preservation when large-scale training is infeasible. However, we find that noise initialization alone is insufficient to capture the detailed motion and content for video stylization.

Gaussian noise is widely adopted as a standard initialization for generative models \cite{ddpm, flow_matching}. Nevertheless, as shown in Figure \ref{fig:noise_init}, while the stylized reference image retains the overall semantic layout, it is treated as an \textbf{out-of-distribution} example, and the video generated from the same Gaussian noise fails to exhibit realistic dynamic motion. In contrast, the noise obtained through inversion preserves critical dependencies across pixels \cite{staniszewski2024there}. Thus, we employ inverted noise for video stylization, which enables partial recovery of camera and object motion; however, content not visible in the first frame remains irrecoverable, and the color is inconsistent with the reference. Therefore, we are seeking additional strategies for more accurate motion and layout preservation in the stylized video.

We further observe that the generated video content is highly sensitive to the reference, and replacing this frame with an edited or stylized one after inversion often introduces significant artifacts. We attribute this issue to the strong attention coupling between the noise and reference latents during the entire denoising stages (Figure \ref{fig:attention_vis}). Specifically, the noise latent estimated through inversion tends to overfit the distribution of the original reference image. To mitigate this issue, we propose to remove the reference image during inversion, which substantially reduces these artifacts.

\begin{figure}[!t]
    \vspace{-1ex}
    \includegraphics[width=\linewidth]{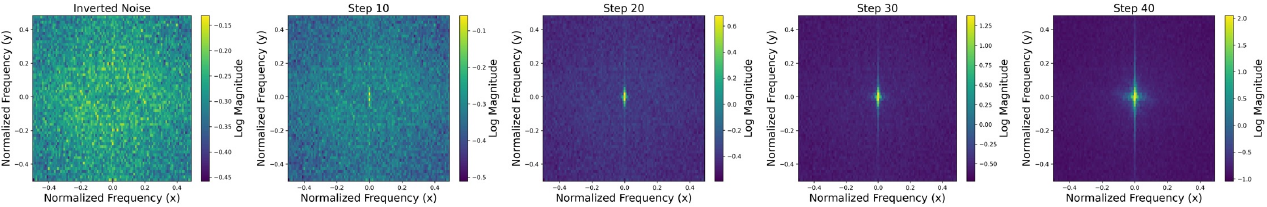}
    \vspace{-4ex}
    \caption{
    \textbf{Frequency spectrum of intermediate latents across timesteps.} We apply Fast Fourier Transform (FFT) along the two spatial dimensions. Lower frequencies are concentrated near the center, whereas higher frequencies are distributed toward the corners.
    \vspace{-1ex}
}
    \label{fig:spectrum}
\end{figure}
\subsection{Frequency Analysis of Intermediate Latents}
\label{app:frequency}
In the frequency spectrum of the initially inverted noise, all frequency components exhibit similar intensity (Figure \ref{fig:spectrum}), characteristic of Gaussian white noise. As the denoising process progresses, the intensity of high-frequency (HF) components diminishes, while low-frequency (LF) components become more pronounced. This transition occurs most noticeably during the early stages. Therefore, our Indirect High-frequency Compensation (IHC) strategy is only applied in the early timesteps.

\begin{wrapfigure}{l}{0.5\textwidth}  
\vspace{-2ex}  
\includegraphics[width=\linewidth]{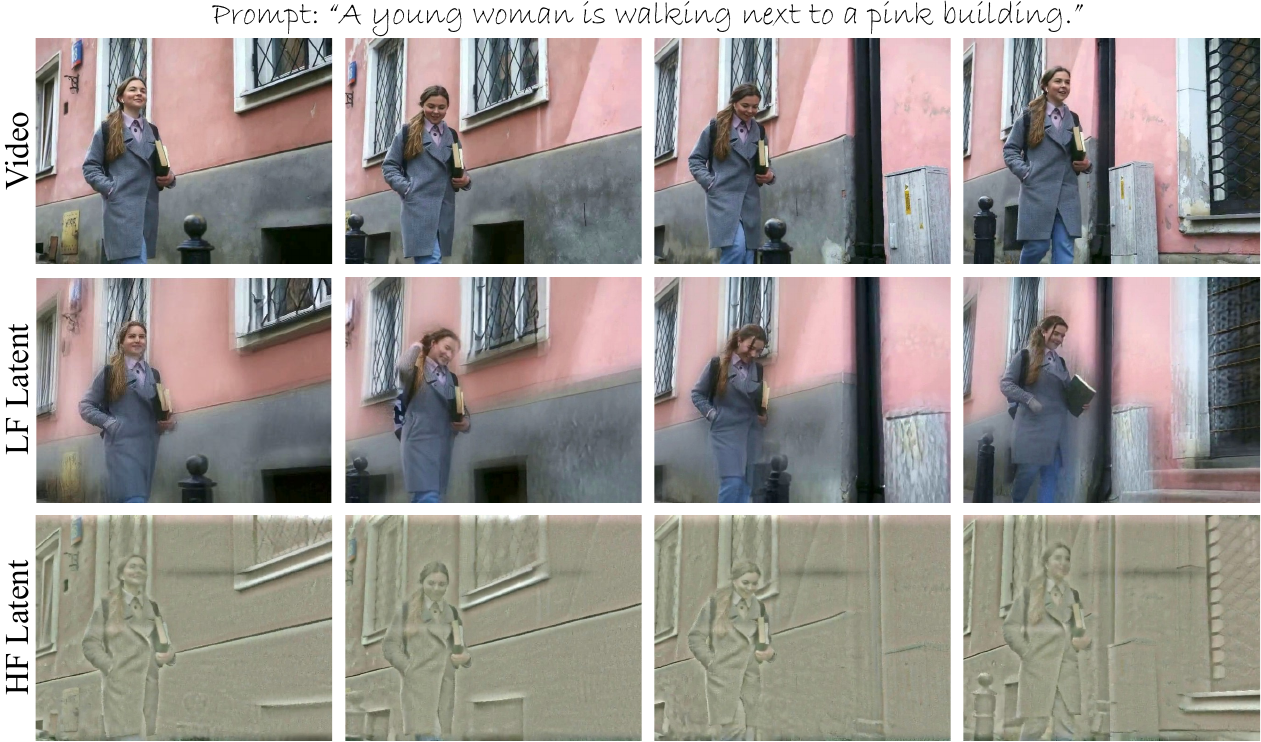}
\caption{
Effects on video reconstruction of LF and HF compensations in PnP inversion. 
}
\label{fig:frequency}
\vspace{-2ex}  
\end{wrapfigure}
We observe that different frequency components of the intermediate latents serve distinct roles when applying PnP inversion during inference to enhance content preservation. As Figure \ref{fig:frequency} illustrates, the low-frequency (LF) components primarily govern the appearance attributes, such as the color distribution, of the generated video. On the contrary, the high-frequency (HF) components predominantly encode layout and motion information. Consistent with characteristics observed in natural images, HF components capture edge and boundary details, which are critical for determining the structural layout. Since the HF component contains less color information, which will not affect the stylized results, we leverage it to constrain the layouts of the stylized videos in our method.

\subsection{Limitations \& Future Works}
\textbf{Time Complexity.} Similar to other training-free image \& video editing approaches, FreeViS requires an inversion process to obtain the denoising trajectory, which usually doubles the inference time. In the meanwhile, since FlashAttention~\cite{dao2024flashattention} does not provide optimized support for masked attention, the use of masked attention in Section~\ref{sec:method} increases the overall time complexity. In practice, FreeViS adds approximately 30\% overhead to the stylization time of the AnyV2V* implementation. A straightforward workaround is to use FreeViS for generating original–stylized video pairs prior to the supervised fine-tuning (SFT) of a video diffusion model, thereby compensating for the scarcity of such paired data. Combining with recent advances of auto-regressive video diffusion models, real-time video stylization can be possible.

\begin{figure}[!t]
    \vspace{-1ex}
    \includegraphics[width=\linewidth]{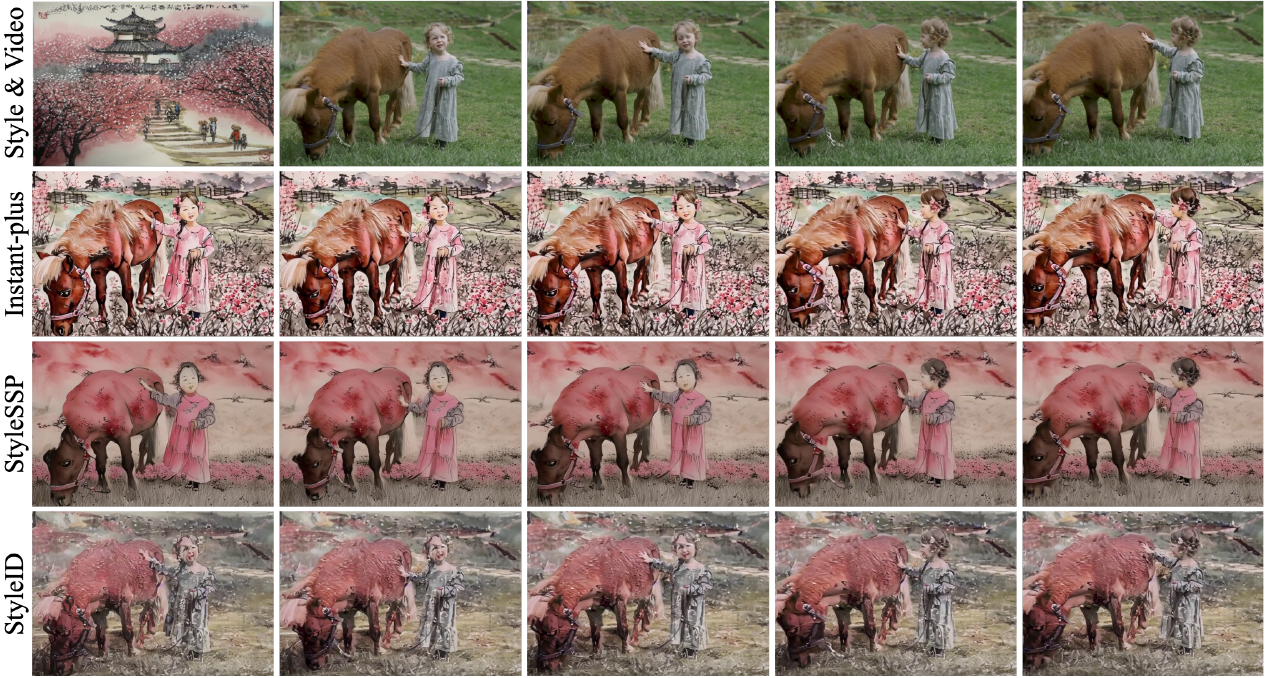}
    \vspace{-4ex}
    \caption{
    Video stylization results for FreeViS combined with different image style transfer models. "Instant-plus" denotes InstantStyle-plus \cite{wang2024instantstyle_plus}.
    \vspace{-1ex}
}
    \label{fig:style_method}
\end{figure}

\textbf{Stylization Upper Bound.} The stylization quality of the references is fundamentally constrained by the underlying image style transfer models. FreeViS enforces the pretrained base model to propagate style features from the reference frames across the entire video. However, the base model itself lacks the capability to explicitly parse or interpret style features. As a result, the overall quality of video stylization, both in terms of style similarity and content preservation, is fully determined by the chosen image model. As illustrated in Figures~\ref{fig:style_method} and~\ref{fig:style_method2}, different image-based models exhibit distinct stylization biases. Future work could explore a mixture-of-experts strategy to selectively combine their complementary strengths according to the style image or video content.

\textbf{Optical Flow Estimation.} FreeViS leverages optical flow to identify overlapping regions between reference frames in order to resolve temporal conflicts. However, conventional optical flow estimation methods, such as RAFT~\cite{teed2020raft} used in this work, struggle with severe occlusion, leading to temporal inconsistencies in the affected regions. More advanced flow estimation and occlusion-handling techniques could be employed to mitigate this issue. Alternatively, dense and precise feature correspondences extracted with DINOv3~\cite{simeoni2025dinov3} offer a promising direction for constructing more reliable inconsistency masks.

\subsection{FreeViS with Various Image Stylization Methods}
\label{app:other_image_stylization}
Leveraging the strong prior of the diffusion model to accomplish high-quality image stylization has become mainstream in recent works \cite{chung2024style, xu2025stylessp, wang2024instantstyle}. FreeViS utilizes the high stylization quality of these methods to accomplish video stylization. In this section, we show that different image stylization methods have different strengths, and FreeViS can be generalized to various methods and maintain their strengths. Three state-of-the-art image style transfer models are selected for experiments: StyleID \cite{chung2024style}, StyleSSP \cite{xu2025stylessp}, and InstantStyle-plus \cite{wang2024instantstyle_plus}. 

\begin{figure}[!t]
    \vspace{-1ex}
    \includegraphics[width=\linewidth]{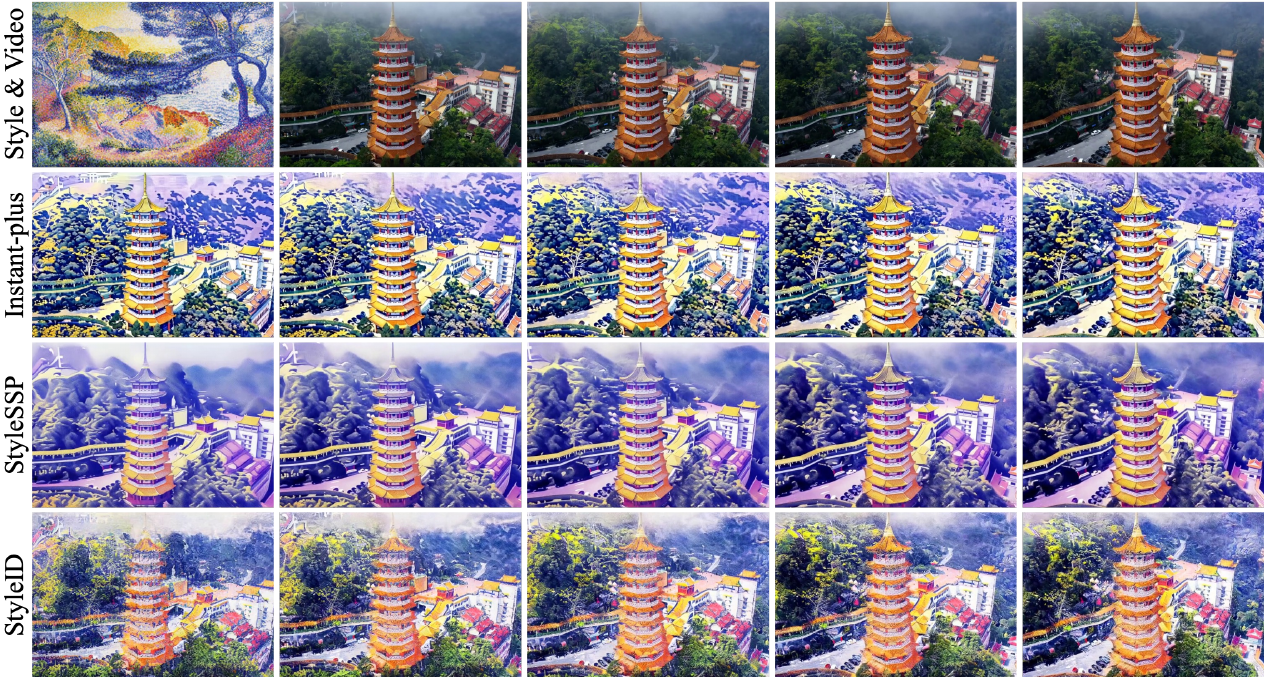}
    \vspace{-4ex}
    \caption{
    Video stylization results for FreeViS combined with different image style transfer models. "Instant-plus" denotes InstantStyle-plus \cite{wang2024instantstyle_plus}.
    \vspace{-1ex}
}
    \label{fig:style_method2}
\end{figure}

As shown in Figures~\ref{fig:style_method} and~\ref{fig:style_method2}, InstantStyle-plus and StyleSSP achieve stronger high-level style alignment, whereas StyleID emphasizes low-level texture preservation. Among them, StyleSSP produces smoother object surfaces and better layout reconstruction compared to the other two methods. InstantStyle-plus provides a more favorable trade-off between visual plausibility and style preservation. Since FreeViS is compatible with all these methods, the underlying image style transfer model can be flexibly substituted to meet the requirements of different scenarios.

\subsection{More Results on Video Stylization}
\begin{figure}[!t]
    \vspace{-1ex}
    \includegraphics[width=\linewidth]{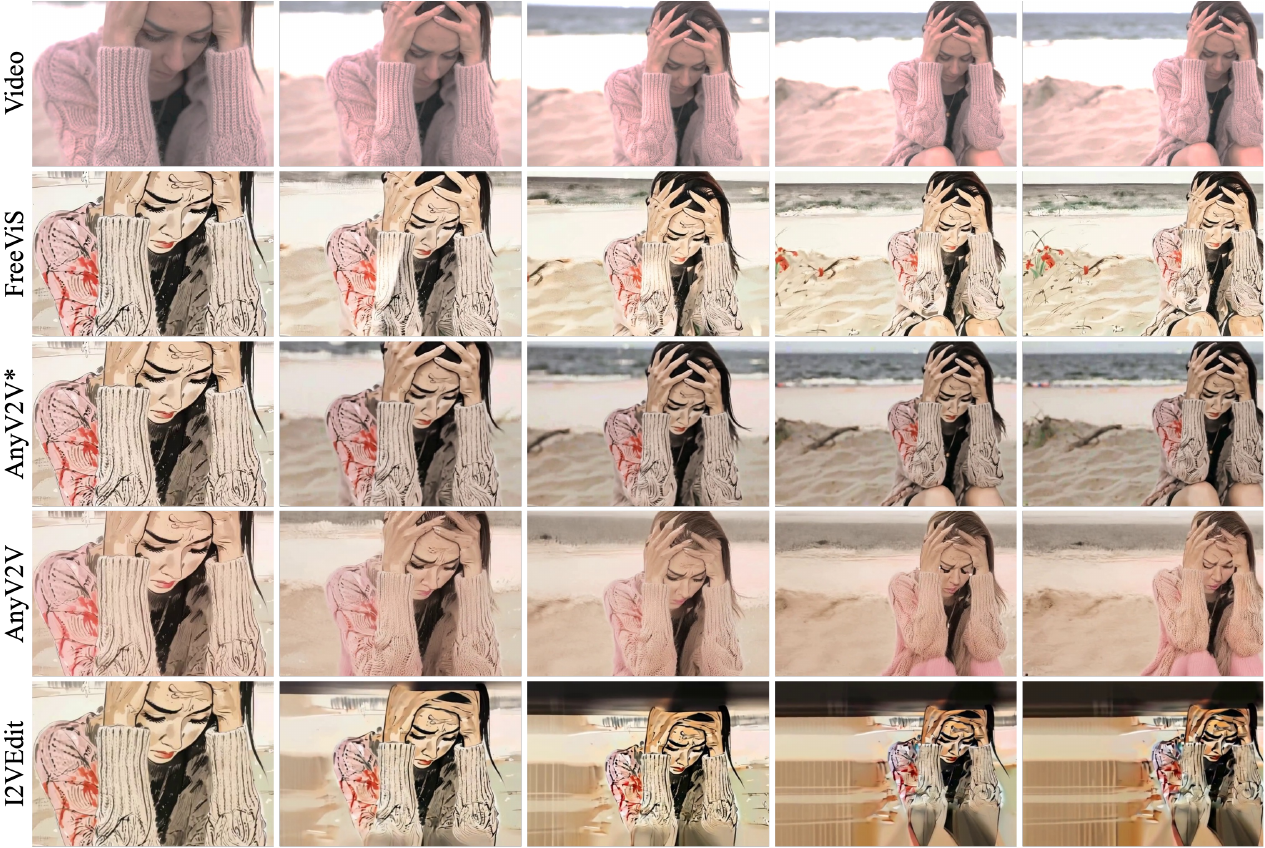}
    \vspace{-4ex}
    \caption{
    Qualitative comparison of FreeViS with other reference-based video editing methods. All the methods are provided with the same first reference.
    \vspace{-1ex}
}
    \label{fig:more_vs1}
\end{figure}

\begin{figure}[!t]
    \vspace{-1ex}
    \includegraphics[width=\linewidth]{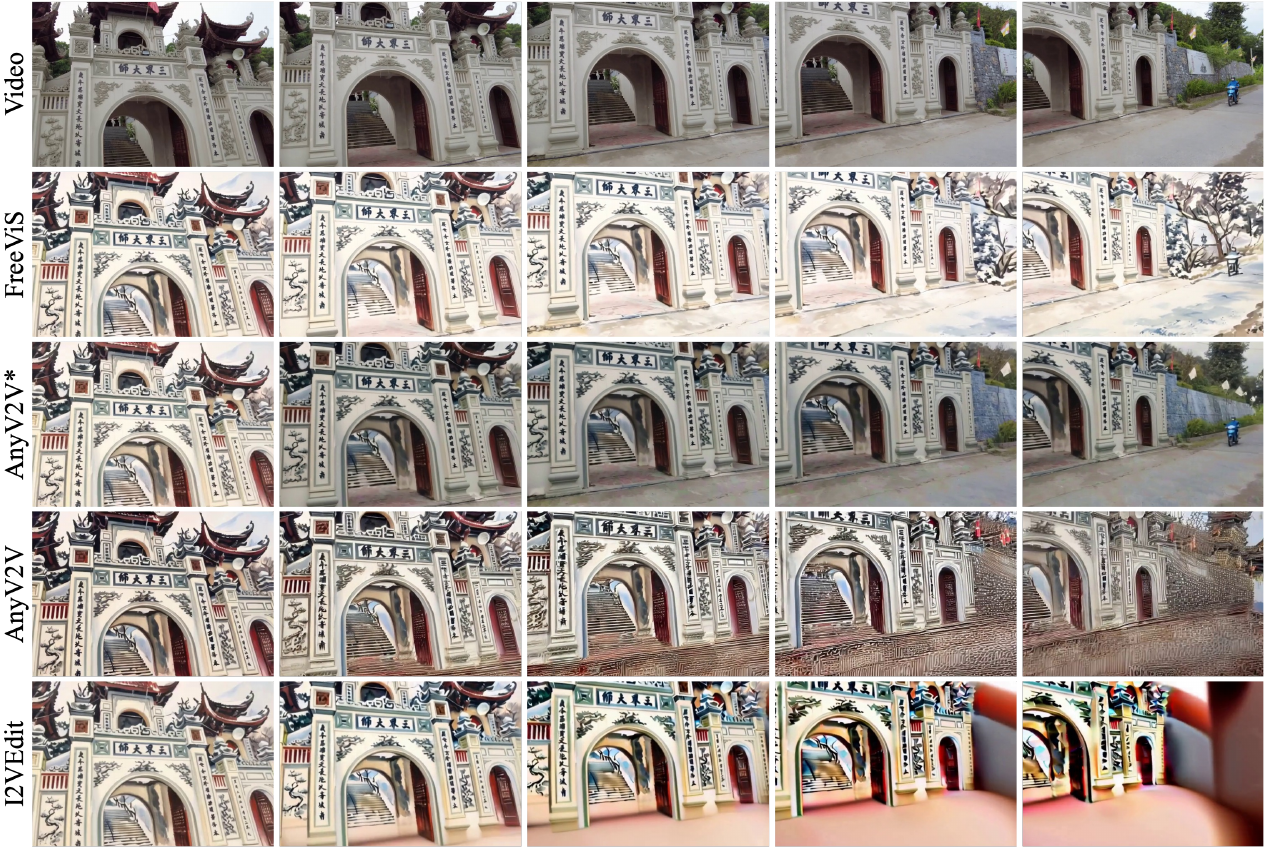}
    \vspace{-4ex}
    \caption{
    Qualitative comparison of FreeViS with other reference-based video editing methods. All the methods are provided with the same first reference.
    \vspace{-1ex}
}
    \label{fig:more_vs2}
\end{figure}

In the main paper, we omit the results of AnyV2V~\cite{ku2024anyv2v} and I2VEdit~\cite{ouyang2024i2vedit}. For completeness, their results are provided in Figures~\ref{fig:more_vs1} and~\ref{fig:more_vs2}. Both methods fail to faithfully reconstruct the essential layout of the original video and exhibit the previously discussed propagation errors, primarily due to their reliance on a single reference input. AnyV2V* (our re-implementation of AnyV2V on the Wan model) demonstrates improved layout preservation. In contrast, FreeViS effectively addresses the propagation error and achieves the highest stylization quality.
\label{app:more_v2v}

\subsection{More Results on Stylized T2V Generation}
\label{app:other_base_model}
\begin{figure}[!t]
    \vspace{-1ex}
    \includegraphics[width=\linewidth]{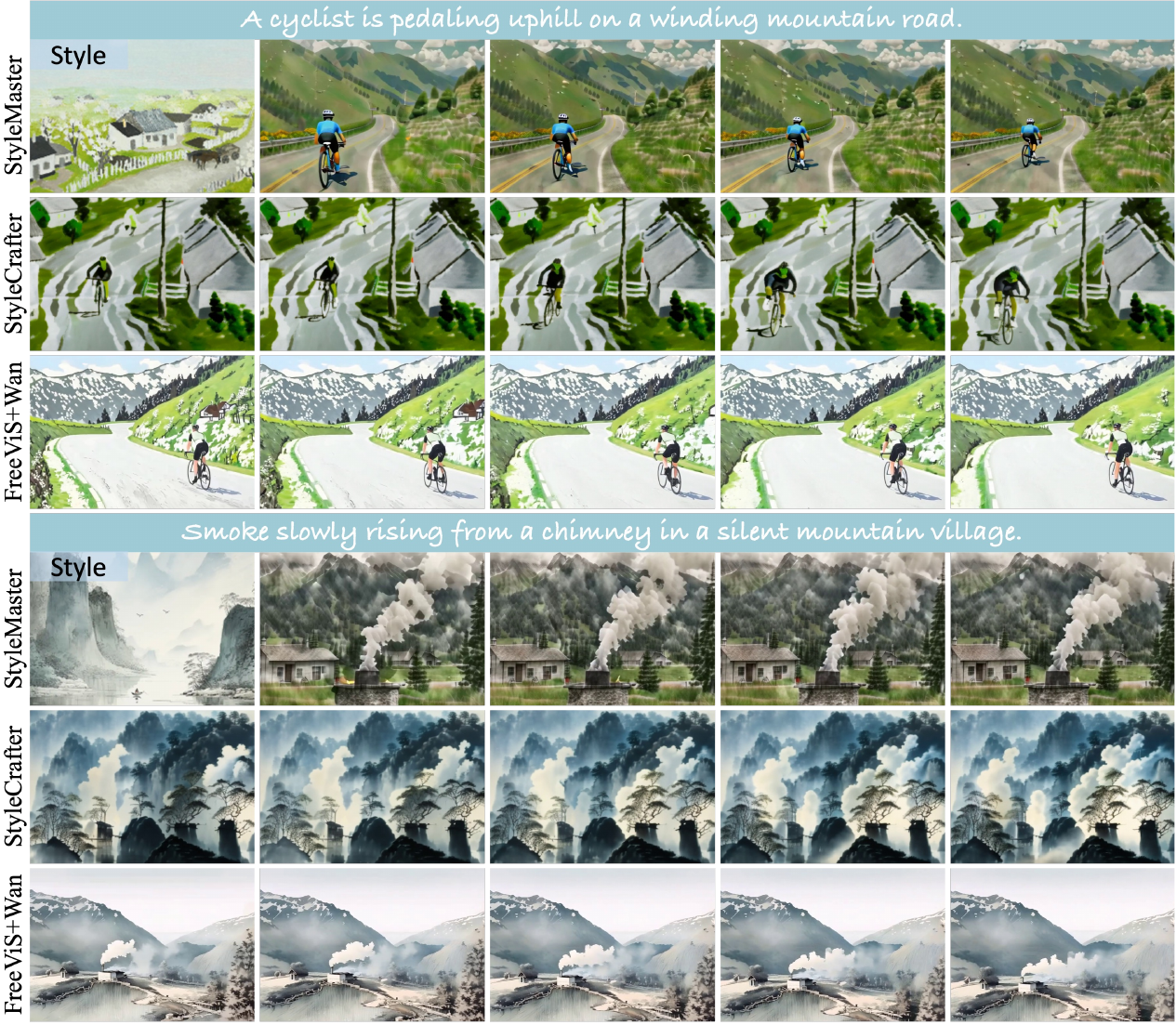}
    \vspace{-4ex}
    \caption{
    Qualitative comparison of FreeViS with other stylized video generation frameworks.
    \vspace{-1ex}
}
    \label{fig:more_t2v}
\end{figure}

Additional results on stylized video generation are shown in Figure~\ref{fig:more_t2v}. Consistent with previous observations, StyleCrafter~\cite{liu2024stylecrafter} demonstrates strong style preservation but struggles with prompt fidelity and motion diversity. In contrast, StyleMaster~\cite{ye2025stylemaster} achieves better text–video alignment but fails to effectively transfer the target style. The combination of FreeViS with the Wan model provides the best balance between style preservation and prompt fidelity, yielding results that are also more visually plausible.
\label{app:more_t2v}

\subsection{FreeViS with Various Video Diffusion Models}
\label{app:other_video_diffusion}
\begin{figure}[!t]
    \vspace{-1ex}
    \includegraphics[width=\linewidth]{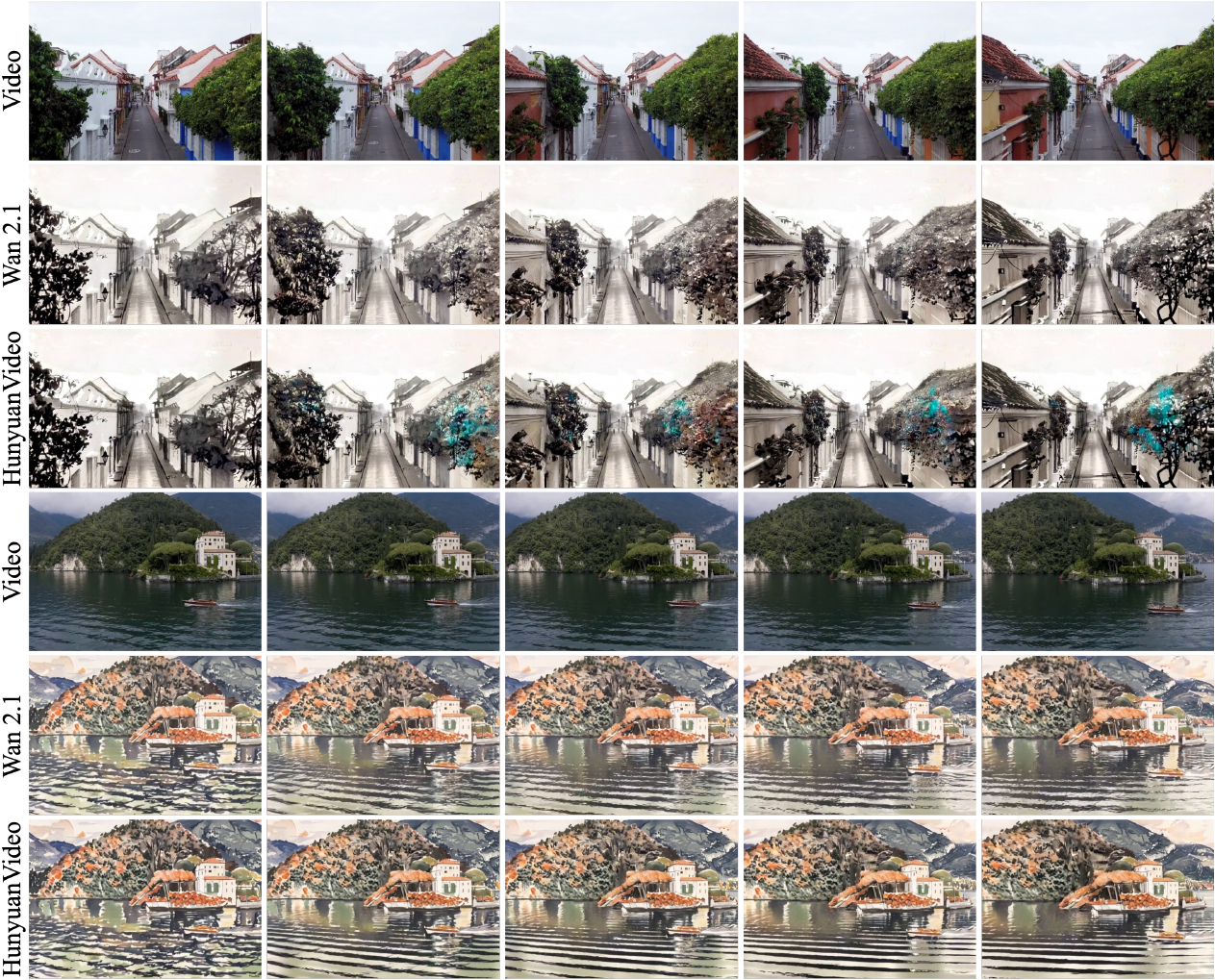}
    \vspace{-4ex}
    \caption{
    Qualitative comparison of FreeViS combined with different base models: Wan \cite{wan2025wan} and HuanyuanVideo \cite{kong2024hunyuanvideo}. Their stylization results in most cases are similar, while Wan presents smoother reconstruction and better temporal consistency. The third row shows a failure case of HuanyuanVideo in consistency preservation.
    \vspace{-1ex}
}
    \label{fig:hunyuan}
\end{figure}
To further validate the compatibility of FreeViS, we implement it on another pretrained I2V model, HunyuanVideo-I2V~\cite{kong2024hunyuanvideo}. Owing to its token replacement mechanism, the first reference latent is tightly coupled with the latents of subsequent frames, which leads to severe overfitting during inversion. As a result, standard inversion fails to reconstruct the video and introduces substantial artifacts. In contrast, PnP inversion with compensation proves both more effective and essential, ensuring faithful video reconstruction. We present a qualitative comparison of FreeViS implemented on Wan~\cite{wan2025wan} and HunyuanVideo in Figure~\ref{fig:hunyuan}. In most cases, the two models achieve comparable stylization performance, while the Wan model demonstrates superior robustness and temporal consistency. These results highlight the importance of leveraging the strong prior of an advanced video diffusion model to achieve higher stylization quality with FreeViS.

\end{document}